\pdfoutput=1
\documentclass[11pt]{article}

\usepackage[final]{acl}

\usepackage{times}
\usepackage{latexsym}
\usepackage[T1]{fontenc}
\usepackage[utf8]{inputenc}
\usepackage{microtype}
\usepackage{inconsolata}
\usepackage{graphicx}
\graphicspath{{figures/}}
\usepackage{booktabs}
\usepackage{multirow}
\usepackage{makecell}
\usepackage{xcolor}
\usepackage{colortbl}
\definecolor{halthl}{HTML}{FCE7E4}
\definecolor{hdrgray}{HTML}{E8E8E8}
\usepackage{amsmath}
\usepackage{amssymb}
\usepackage{xspace}
\usepackage{url}
\usepackage{placeins}
\usepackage{float}

\newcommand{\halt}{HALT\xspace}
\newcommand{\full}{\textsc{full}\xspace}
\newcommand{\any}{\textsc{any\_match}\xspace}
\newcommand{\all}{\textsc{all\_match}\xspace}
\newcommand{\pp}{\,pp\xspace}

\title{HALT: Verification-Aware Stopping for\\Retrieval-Augmented Search Agents\thanks{Code is available at \url{https://github.com/Noverse0/HALT}.}}

\author{Daeyoung Roh \\
  Independent Researcher \\
  \texttt{dybroh@gmail.com} \\
  \And
  Donghee Han \\
  KAIST \\
  \texttt{venzino.han@gmail.com} \\}

\begin{document}
\maketitle

\begin{abstract}
Retrieval-augmented search agents answer multi-hop questions by repeatedly
issuing search queries and accumulating evidence. This creates a stopping
problem: after the necessary evidence has appeared, further retrieval often adds
cost, latency, and distracting context rather than useful information. We frame
stopping as evidence coverage rather than generator confidence, and introduce
\halt, a lightweight verification-aware policy that leaves the search agent
unchanged. Given expected hop claims, \halt{} stops only when cumulative evidence
supports each required claim. Across three multi-hop QA benchmarks, \halt{}
reduces redundant search while largely preserving exact match. We separate a
deployable setting, where hop claims are generated from the question, from a
diagnostic upper bound that uses gold supporting-fact annotations: generated
claims give smaller but still exact-match-preserving savings, while gold claims
show the larger savings available when hop targets are clean.
Baseline comparisons and ablations show that this behavior is driven by
claim--evidence alignment rather than generic sufficiency, fixed stop positions,
or lexical overlap. Open-corpus pilots further suggest that \halt{} abstains when
coverage cannot be reliably verified. Overall, evidence coverage provides a
practical runtime control signal for improving retrieval-augmented agents without
retraining or modifying the host agent.
\end{abstract}

\section{Introduction}
\label{sec:intro}

\begin{figure*}[t]
\centering
\includegraphics[width=\textwidth]{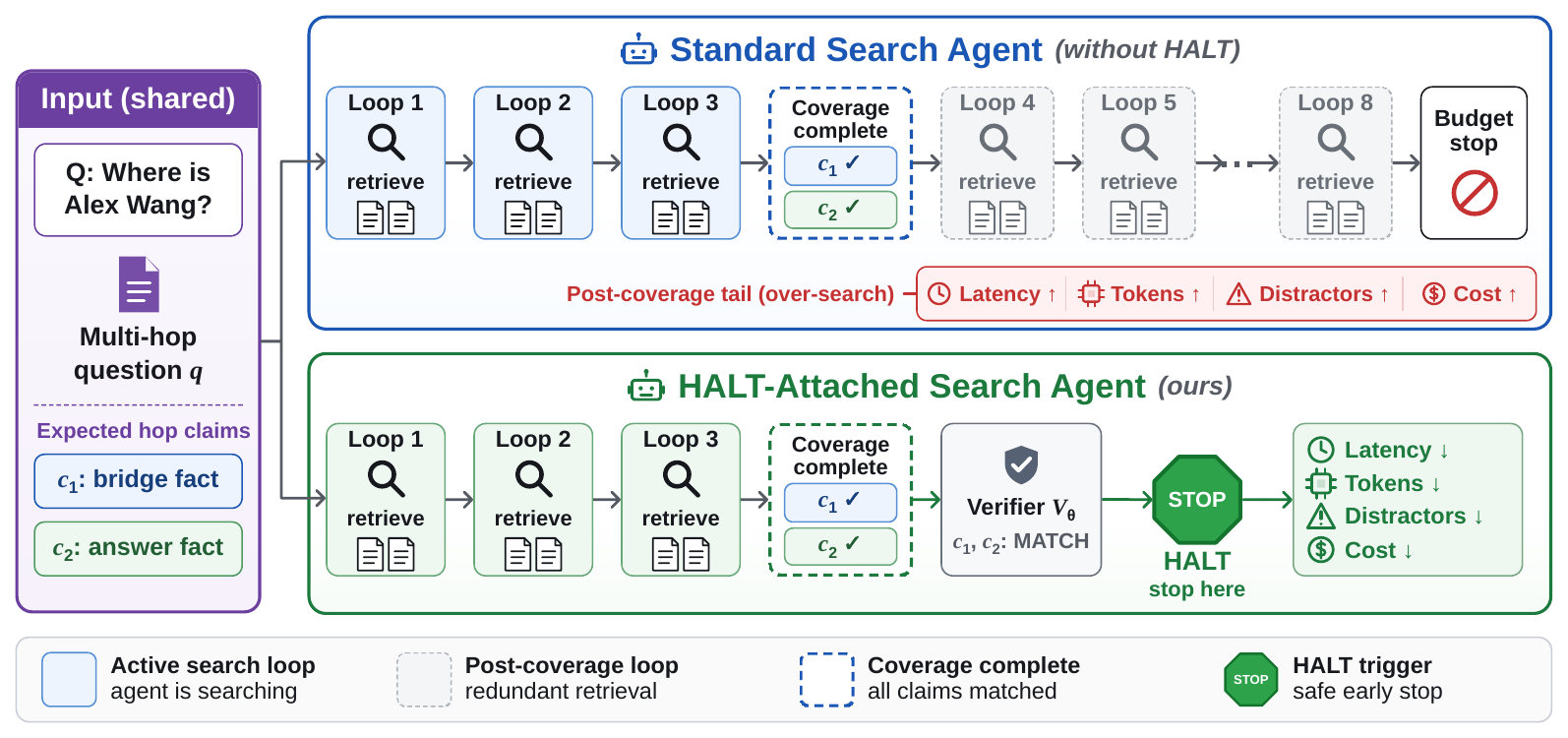}
\caption{\textbf{Post-coverage search.}
A standard agent may keep retrieving after all expected hop claims are
covered. \halt stops once a verifier matches every claim to the
cumulative evidence.}
\label{fig:motivation}
\end{figure*}

Retrieval-augmented generation (RAG) systems ground language models in
external evidence. In agentic RAG, this retrieval step becomes an
iterative search process: the model generates queries, retrieves
evidence, and conditions subsequent queries on the evolving trajectory.
We refer to such systems as retrieval-augmented search agents. This
multi-step search is useful. Later queries can recover bridge entities
or fill missing intermediate evidence, but the same loop creates a
stopping problem. After the needed evidence has appeared, the agent may
keep searching; in our setting, a 3B Self-Ask agent runs to its search
budget on 78--88\% of questions instead of halting on its own. Those
extra calls add latency and cost, and may
introduce irrelevant context that makes the final answer harder rather
than easier (Figure~\ref{fig:motivation}).

Retrieval-augmented search agents include self-asking decomposition
agents \citep{press2023selfask}, ReAct-style tool use
\citep{yao2023react}, and reinforcement-learned search policies
\citep{jin2025searchr1,chen2025research}. In these systems, stopping is
not a cosmetic implementation choice. It determines how much evidence
enters the context, how much computation is spent, and when the agent
hands control back to the answer generator.

Most adaptive-retrieval and stopping methods decide this using
generator-side or trajectory-level signals. They inspect answer
confidence, reflection tokens, hidden states, trajectory state, or
token-level uncertainty to decide whether another retrieval call is
needed
\citep{jiang2023active,trivedi2023interleaving,asai2024selfrag,su2024dragin}.
These signals are useful, but they do not ask the question we need for
multi-hop stopping. A model can be confident from partial evidence. It
can also keep searching even after the retrieved evidence already covers
the necessary steps. We instead ask a retrieval-side question: has the
agent retrieved evidence for every reasoning hop required by the
question?

We introduce \textbf{\halt} (\textbf{H}op-\textbf{A}ligned
\textbf{L}oop \textbf{T}ermination), a verification-aware stopping
policy for multi-hop search agents. \halt leaves the host search agent
unchanged. After each retrieval loop, a small supervised verifier checks
the cumulative retrieved evidence against a set of expected hop claims.
For each claim, the verifier predicts \textsc{match}, \textsc{partial},
or \textsc{null}. The agent stops only when all expected claims are
matched. Thus, \halt does not score the trajectory or candidate answer
as globally sufficient; it asks whether each required hop has evidence.

This design keeps the stopping predicate separate from the search
policy. The search agent is not retrained, the retriever is unchanged,
and the generator need not expose an internal confidence signal. The
predicate is also explicitly per-hop. Matching one relevant fact is not
enough if another required hop remains unsupported. In our experiments,
the hop claims are obtained in two ways: from supporting-fact annotations
for diagnostic experiments, or generated from the question alone when
such metadata is unavailable.

We evaluate \halt on HotpotQA, 2WikiMultihopQA, and MuSiQue using
Self-Ask agents with 3B and 7B backbones. A verifier trained once on
HotpotQA discrepancy labels is reused across datasets and Self-Ask
backbone scales. In the closed BM25 distractor setting, our deployable
condition generates hop claims from the question alone: savings are
smaller and dataset-dependent, but paired non-inferiority tests show
that exact match (EM) is preserved. Gold supporting-fact claims serve as
a diagnostic upper bound, showing the larger savings available when hop
targets are clean---most strongly on the 3B agent, and with EM preserved
in all 7B gold-claim cells. In open-corpus pilots, \halt becomes cautious: it
fires on only a small subset of HotpotQA questions without EM loss, and
on only $2.4\%$ of Bamboogle questions, again at no EM cost. We
therefore treat open-corpus retrieval
as a stress test, not as the main efficiency setting.

Our contributions are:
\begin{itemize}
\item \textbf{Stopping as evidence coverage.}
We formulate stopping in retrieval-augmented search agents as a test of
whether retrieved evidence covers the reasoning hops required by the
question, rather than whether the generator appears ready to answer.

\item \textbf{Hop-aligned loop termination.}
We introduce \halt, which adds a small per-hop evidence verifier to a
frozen search agent and changes only the halting predicate, leaving the
search policy, retriever, and generator unchanged.

\item \textbf{A distinct stopping signal.}
On identical trajectories, \halt reduces search loops while preserving
standardized-extractor EM in the main controlled setting. Baseline
comparisons and ablations show that its decisions are not explained by
generic sufficiency, fixed stop positions, expected-page matching, or
surface-form overlap.
\end{itemize}

\section{Background and Related Work}
\label{sec:related}

\paragraph{Retrieval-augmented search agents.}
Retrieval-augmented generation (RAG) grounds language models in external
evidence \citep{lewis2020rag}. In agentic RAG, retrieval is no longer a
single preprocessing step: the model can issue queries over multiple
turns and use the evolving trajectory to decide what to search next.
Self-Ask \citep{press2023selfask} decomposes a question into follow-up
sub-questions, ReAct \citep{yao2023react} interleaves reasoning with
tool use, and recent systems such as Search-R1
\citep{jin2025searchr1} and ReSearch \citep{chen2025research} learn
search policies with reinforcement learning. Such policies can internalize
termination through the training reward \citep{java2025frugalrag}, yet
recent analyses find that even reinforcement-learned search agents
frequently over-search or under-search relative to the evidence actually
required \citep{wu2025hiprag,wu2025searchwisely}. These systems make search
adaptive, but termination is usually left to the agent policy or the
generation process. We study it as a separate decision: whether the
evidence retrieved so far covers the reasoning hops required by the
question.

\paragraph{Adaptive retrieval.}
Adaptive-retrieval methods decide when generation should call a
retriever. FLARE \citep{jiang2023active} retrieves from low-confidence
tokens, while DRAGIN \citep{su2024dragin} estimates token-level
information need with its RIND signal. These methods are retrieval
triggers inside decoding. HALT addresses a different point in the
pipeline. The search loop is already active; the question is when its
accumulated evidence is enough to stop. We include adaptive-retrieval
methods as retrieval-saving controls, but in multi-hop QA reducing calls
is useful only if the stopped trajectory still preserves answer
accuracy. More broadly, when-to-retrieve methods range from
self-knowledge routing \citep{wang2023skr} to uncertainty-based triggers,
and a recent benchmark argues that simple uncertainty estimates can match
more elaborate adaptive pipelines on efficiency
\citep{moskvoretskii2025adaptive}. We therefore report retrieval cost
alongside accuracy.

\paragraph{Adaptive stopping and stopping signals.}
The closest work asks when an iterative retrieval loop should halt, but
most stopping signals are global. Self-RAG \citep{asai2024selfrag} uses
reflection-token critics; Stop-RAG \citep{park2025stop} learns a
value controller over trajectory state; SIM-RAG
\citep{yang2025simrag} scores the sufficiency of a
(question, evidence, intermediate answer) tuple; SeaKR
\citep{yao2024seakr} measures the model's internal-state uncertainty to
decide when to re-retrieve; and IRCoT
\citep{trivedi2023interleaving} stops when the reasoning chain produces an
answer span. A common thread is that these signals estimate generator
confidence, internal-state uncertainty, or global answer sufficiency.
Other controls operate at coarser or later stages:
Adaptive-RAG \citep{jeong2024adaptiverag} selects a retrieval depth from
question complexity, while SuRe \citep{kim2024sure} verifies candidate
answers after retrieval rather than controlling the loop itself. The
construct closest to this coverage view is the ``sufficient context''
notion of \citet{joren2025sufficient}, but it is used to decide
abstention and diagnose hallucination rather than to halt an iterative
search loop. \halt{} differs in what it asks the stopping module to judge: it
scores cumulative retrieved evidence against \emph{per-hop expected
claims} during the loop. Its stopping predicate asks whether each
required hop is covered, not whether the current trajectory or candidate
answer is globally sufficient. This lets the stopping decision
distinguish a trajectory that contains one strong fact from one that
covers the full multi-hop evidence structure.

\paragraph{Multi-hop QA as a stopping testbed.}
HotpotQA \citep{yang2018hotpotqa}, 2WikiMultihopQA
\citep{ho2020constructing}, and MuSiQue \citep{trivedi2022musique}
require evidence from multiple supporting facts or reasoning steps. This
makes them useful for testing stopping policies. Stopping early can miss
a required hop; continuing after coverage adds retrieval cost and can
bring in distracting evidence.
\section{Method}
\label{sec:method}

\begin{figure*}[!t]
\centering
\includegraphics[width=\textwidth]{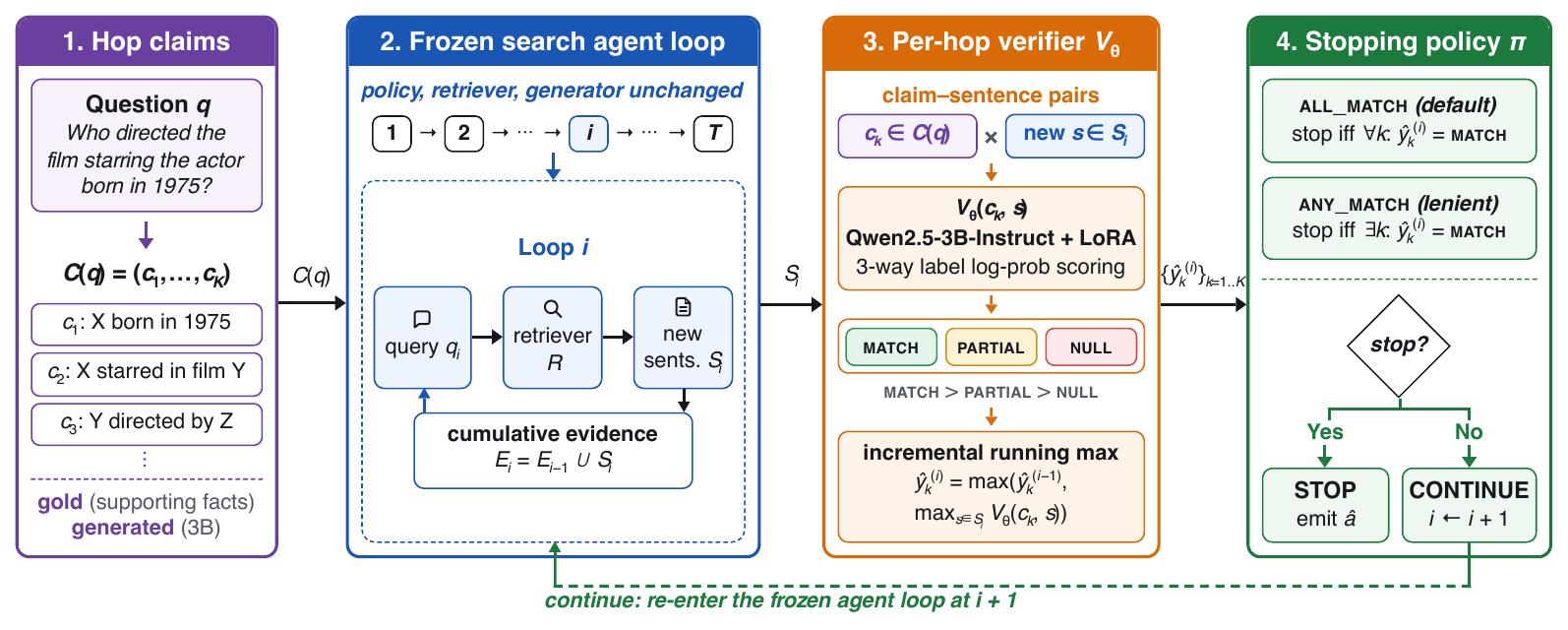}
\caption{\textbf{HALT pipeline.} HALT adds a verification-aware
stopping gate on top of a frozen search agent --- policy, retriever,
and generator are untouched. Expected hop claims $C(q)$ (gold
supporting facts, or generated zero-shot from the question) are scored
against each loop's newly retrieved sentences by the per-hop verifier
$V_\theta$, which keeps an incremental running-max label per claim; the
stopping policy $\pi$ (default \all: every claim \textsc{match}) halts
at the first satisfying loop, otherwise the agent re-enters loop
$i{+}1$.}
\label{fig:method-overview}
\end{figure*}

\subsection{Verification-aware Stopping}
\label{sec:verification-aware-stopping}

Let a search trajectory be a sequence of retrieval loops
$\tau=(l_1,l_2,\ldots,l_T)$. At loop $l_i$, the agent retrieves a set of
sentences $S_i$, and the cumulative evidence is
$E_i=\bigcup_{j \le i} S_j$. For a multi-hop question $q$, HALT assumes
expected hop claims $C(q)=(c_1,c_2,\ldots,c_K)$, where each claim states
the evidence needed for one reasoning hop
(\S\ref{sec:claim-construction}).

A verifier $V_\theta$ assigns each claim--sentence pair one of
$\{\textsc{match},\textsc{partial},\textsc{null}\}$. For each claim,
HALT keeps the strongest label observed so far over all retrieved
sentences, ordered as \textsc{match} $>$ \textsc{partial} $>$
\textsc{null}. Since evidence only accumulates, each loop scores only
the newly retrieved sentences $S_i$ and updates these running labels.

Let $\hat{\mathbf{y}}^{(i)}=(\hat{y}^{(i)}_1,\ldots,\hat{y}^{(i)}_K)$
be the claim-level labels after loop $i$. A stopping policy $\pi$ maps
this vector to a halting decision, and the agent stops at the first loop
$i$ for which $\pi(\hat{\mathbf{y}}^{(i)})$ is true. We evaluate:
\begin{itemize}
\item \any: stop when at least one expected hop claim is
\textsc{match}.
\item \all: stop only when every expected hop claim is
\textsc{match}.
\end{itemize}

\all is the default policy. \any is a permissive contrast that tests
whether matching one hop is enough, rather than complete hop coverage.

\subsection{Expected Hop Claim Construction}
\label{sec:claim-construction}

HALT needs a target claim for each expected reasoning hop. We use two
sources.

\textbf{Gold claims} are constructed from benchmark supporting-fact
annotations. They provide clean hop targets and isolate the stopping
mechanism from claim-generation error, but rely on metadata unavailable
in deployment. We use them as a diagnostic condition.

\textbf{Generated claims} are produced zero-shot from the question alone
using Qwen2.5-3B. This removes supporting-fact metadata and tests
whether the same verifier can stop from predicted hop targets. Both
claim sources use the same verifier input schema. Appendix~\ref{sec:appendix-claim-quality}
reports the record format, prompt, and generated--gold alignment
($F_1{=}0.701$ on 2Wiki and $0.455$ on HotpotQA).

\subsection{Verifier Architecture and Training}
\label{sec:verifier}

We instantiate $V_\theta$ with Qwen2.5-3B-Instruct
\citep{qwen2_5} and train a LoRA adapter \citep{hu2022lora} by
supervised fine-tuning the causal language model to emit the label of a
claim--sentence pair; no separate classification head is added. At
inference, the verifier scores each of the three label strings
\textsc{match}, \textsc{partial}, and \textsc{null} by its
length-normalized token log-probability and predicts the argmax.
Training examples are built automatically
from HotpotQA: supporting-fact annotations provide claim--evidence pairs
for required hops, and distractor passages provide unsupported pairs.
The discrepancy-label procedure in \S\ref{sec:appendix-verifier} maps
examples to \textsc{match}, \textsc{partial}, or \textsc{null}.

The balanced training set contains $9{,}626$ rows from $4{,}903$
questions. We train for $1{,}000$ steps and obtain $0.9304$ macro-F1 on
a question-disjoint dev split of $2{,}374$ rows. The verifier is trained
once on HotpotQA and reused without tuning across Self-Ask backbones
(Qwen2.5-3B/7B) and the three benchmark datasets.
Additional verifier diagnostics are given in
\S\ref{sec:appendix-verifier}.

\subsection{Statistical Testing Protocol}
\label{sec:stats}

For each dataset--backbone cell, we evaluate standardized EM and average search loops. EM is computed by a frozen
Qwen2.5-3B-Instruct extractor applied to the cumulative chunks available
at the stop loop. We define
$\Delta_{\text{EM}}=\text{EM}_{\halt}-\text{EM}_{\full}$ and
$\Delta_{\text{loops}}=\text{loops}_{\halt}-\text{loops}_{\full}$, so
negative $\Delta_{\text{loops}}$ means fewer loops.

We compare \halt \all against \full, which runs the agent to the full
trajectory budget. For each cell, we use a paired bootstrap test
\citep{koehn2004statistical} with $1000$ resamples:
\begin{align}
H_0^{\text{EM}} :\;&  \Delta_{\text{EM}} \le -\varepsilon  \quad (\text{NI}) \\
H_0^{\text{loops}} :\;&  \Delta_{\text{loops}} \ge 0 \quad (\text{sup})
\end{align}
with $\varepsilon=0.02$ (2\,pp). A cell passes only if both EM
non-inferiority and loop superiority reject after Holm correction at
$\alpha=0.05$. The 3B family has $m{=}12$ tests
(2 endpoints $\times$ 3 datasets $\times$ 2 claim sources), and the 7B
family has $m{=}6$ tests (2 endpoints $\times$ 3 datasets, gold claims
only). Because EM is a paired binary endpoint, we also report
paired-binary robustness checks using Newcombe hybrid-score confidence
intervals \citep{newcombe1998paired} and exact McNemar tests
\citep{mcnemar1947note} in \S\ref{sec:appendix-stats}. Margin
justification and per-cell statistics are reported in
\S\ref{sec:appendix-stats}.

\section{Experimental Setup}
\label{sec:setup}

\begin{table*}[t!]
\centering
\small
\begin{tabular}{lrrrrrrrc}
\toprule
Dataset & Full EM & \halt EM & $\Delta$EM & $\Delta$F1 & Full loops & \halt loops & $\Delta$ loops & NI test \\
\midrule
2Wiki    & \underline{0.184} & \textbf{0.199} & $+1.5\,$pp & $+1.3\,$pp & \underline{5.52} & \textbf{3.46} & $-37\%$ & PASS \\
HotpotQA & \textbf{0.306} & \underline{0.297} & $-0.9\,$pp & $-1.5\,$pp & \underline{5.71} & \textbf{3.15} & $-45\%$ & FAIL \\
MuSiQue  & \underline{0.083} & \textbf{0.086} & $+0.3\,$pp & $+0.4\,$pp & \underline{5.56} & \textbf{4.47} & $-20\%$ & PASS \\
\bottomrule
\end{tabular}
\caption{\textbf{Primary diagnostic result: \halt \all on Self-Ask 3B with gold claims.}
EM is from a frozen Qwen2.5-3B extractor; NI uses
$\varepsilon{=}2\,$pp with Holm $m{=}12$ over the full 3B
gold+generated family and records whether non-inferiority was
\emph{demonstrated}. HotpotQA gold is a FAIL: its 95\% CI
$[-2.6,+0.8]\,$pp crosses $-\varepsilon$. MuSiQue uses its $885$ valid
rows ($115$ questions yield no atomic hop claims;
\S\ref{sec:appendix-stats}). $\Delta$F1 tracks $\Delta$EM everywhere,
so the result is not an exact-match artifact.}
\label{tab:primary}
\end{table*}

\paragraph{Datasets and splits.}
We evaluate on three multi-hop QA benchmarks: HotpotQA in the
distractor setting, 2WikiMultihopQA, and MuSiQue. For each dataset, we
draw $n{=}1000$ questions from the standard dev split. The 3B-agent and
7B-agent experiments use the same question subsets. One exception:
$115$ MuSiQue questions yield no atomic gold hop claims, so MuSiQue
gold-claim statistics use $n{=}885$ (\S\ref{sec:appendix-stats}).

\paragraph{Retrieval settings.}
The primary experiments use \emph{Setting~A}, a controlled closed-pool
retrieval setting. For each question, we build a BM25 sentence index
\citep{robertson2009bm25} over the union of gold and distractor
paragraphs supplied by the benchmark. Each sentence is treated as a
candidate chunk, and each follow-up query retrieves top-$K{=}3$
sentences. Because the supporting evidence is present in the candidate
pool, Setting~A isolates the stopping problem from open-corpus
retrieval failure.

We also report a small \emph{Setting~B} pilot with open-corpus
retrieval. The agent retrieves from a global Wikipedia 2018 index
\citep{petroni2021kilt} using dense E5-base retrieval
\citep{wang2022e5}, following the open-corpus Wikipedia setting used by
Search-R1 \citep{jin2025searchr1} and ReSearch
\citep{chen2025research}. Setting~B is closer to deployment, but
retrieval-quality failures can obscure the stopping signal. We treat it
as a pilot rather than the primary statistical testbed; results are in
\S\ref{sec:results-settingB}.

\paragraph{Search agents.}
We use structured agents that expose loop-level retrieval trajectories.
\textbf{Self-Ask} \citep{press2023selfask}, instantiated with
Qwen2.5-3B-Instruct and Qwen2.5-7B-Instruct, is the host agent for all
main results. The maximum trajectory budget is 8 loops; verifier
details are in \S\ref{sec:verifier} and \S\ref{sec:appendix-verifier}.

\paragraph{Baselines.}
We compare \halt with full-budget, adaptive-stopping, and
adaptive-retrieval controls. \textbf{Full} runs the host agent to the
full trajectory budget. The adaptive-stopping baselines are
\textbf{V-StopRAG} \citep{park2025stop}, a value-style classifier
over cumulative evidence and loop index, and \textbf{V-SIMRAG}
\citep{yang2025simrag}, which additionally conditions on the current
sub-question and sub-answer. We also report an \textbf{OR-ensemble} of
\halt and V-StopRAG as a signal-redundancy check.

For adaptive-retrieval controls, we include \textbf{FLARE}
\citep{jiang2023active} and \textbf{DRAGIN-approx}
\citep{su2024dragin}. FLARE triggers retrieval from low-confidence
tokens. DRAGIN's original RIND signal requires attention hooks that are
unavailable in our setup. DRAGIN-approx therefore uses the same
confidence-threshold interface as FLARE, with a higher retrieval
threshold. We treat DRAGIN-approx as a retrieval-control reference
rather than an exact reproduction of DRAGIN. Baseline training details,
thresholds, and hyperparameters are reported in
\S\ref{sec:appendix-repro}.

\paragraph{Answer extraction and EM.}
We report \textbf{Native EM}, computed from each agent's own answer head,
and \textbf{Std-Ext EM}, our primary endpoint, computed by a frozen
Qwen2.5-3B-Instruct extractor over cumulative chunks at the stop loop
with HotpotQA-style normalization. Native EM serves as a faithfulness
check. Unless otherwise noted, all experiments use seed $13$; prompts,
hyperparameters, hardware, and reproducibility details are in
\S\ref{sec:appendix-repro}.
\begin{table}[t]
\centering
\small
\begin{tabular}{lrrrr}
\toprule
Dataset & $\Delta$EM & $\Delta$F1 & $\Delta$ loops & NI test \\
\midrule
2Wiki     & $+0.6\,$pp & $+0.5\,$pp & $-6.3\%$ & PASS \\
HotpotQA  & $-0.5\,$pp & $-0.7\,$pp & $-9.3\%$ & PASS \\
MuSiQue   & $-0.2\,$pp & $-0.1\,$pp & $-6.6\%$ & PASS \\
\bottomrule
\end{tabular}
\caption{\textbf{Cross-scale: \halt on Self-Ask 7B, gold-claim condition.}
The same verifier from \S\ref{sec:results-primary} is used without
retraining; all three cells pass Holm $m{=}6$. $\Delta$F1 tracks
$\Delta$EM on every cell, so the EM result is not an exact-match
artifact.}
\label{tab:scaling}
\end{table}

\section{Results}
\label{sec:results}

\subsection{Primary: \halt at the 3B agent}
\label{sec:results-primary}

Table~\ref{tab:primary} reports the gold-claim diagnostic condition for
the conservative \all policy on the Self-Ask 3B agent. \halt reduces
average search loops on all three datasets while largely preserving
Std-Ext EM: two of three gold cells pass EM non-inferiority, and
HotpotQA gold fails only because its 95\% CI crosses $-\varepsilon$
around a small point decrease ($\Delta\text{EM}{=}{-}0.9\,$pp). Across
the full 3B family, including generated
claims, \halt passes EM non-inferiority in 5/6 cells and loop superiority
in all six; generated claims yield smaller but still EM-preserving loop
reductions. Appendix~\ref{sec:appendix-stats} reports paired-binary
robustness checks, which give the same EM decisions.

The headroom comes from the agent itself: by the logged stop reason,
the 3B agent runs to its search budget instead of halting natively on
$87.5\%$ / $78.1\%$ / $81.2\%$ of questions
(HotpotQA / 2Wiki / MuSiQue), so most trajectories contain
post-coverage search that an external coverage test can remove.

\subsection{Cross-scale transfer to a 7B Self-Ask agent}
\label{sec:results-7b}

We apply the same 3B verifier to a 7B Self-Ask agent without retraining.
Table~\ref{tab:scaling} shows that all three datasets pass both endpoints
under Holm $m{=}6$. Loop savings are smaller than at 3B, consistent with
the 7B agent already stopping earlier. The stop-reason log makes this
concrete: the 7B agent halts natively on essentially every question
(budget exhaustion $0.0$--$0.5\%$), leaving little post-coverage search
to remove. We therefore read the smaller 7B savings as limited headroom
rather than a failure of the coverage signal, which the under-search
analysis below tests directly.

\subsection{Bidirectional control: \halt also detects under-search}
\label{sec:results-continue}

The 3B result shows \halt can safely \emph{cut} search loops. A stronger
agent raises the opposite question: can the same coverage signal detect
when an agent stops \emph{too early}? On the 7B agent, \halt's own \all
criterion is never satisfied at the agent's native stop loop on $51\%$ /
$64\%$ / $79\%$ of questions (HotpotQA / 2Wiki / MuSiQue) --- direct
evidence of under-search at scale, consistent with reports of over- and
under-search in RL search agents \citep{wu2025hiprag,wu2025searchwisely}.
We test whether extending exactly these flagged trajectories recovers EM.

\paragraph{Design.} On the subset of 7B questions that self-stop without
reaching \all ($n{=}511/635/723$, after deduplicating repeated MuSiQue
question ids), we re-run the agent
with a \texttt{--force-continue} flag that overrides the agent's native
stop and continues loop-by-loop until \all fires or the budget (8 loops)
is exhausted; \halt then stops each forced trajectory at its own \all
loop. We disclose that, unlike the cutting direction, this overrides the
agent's own stop decision --- \halt still supplies only the continuation
\emph{signal} (full protocol and extractor-consistency details in
Appendix~\ref{sec:appendix-continue}).

\paragraph{Forcing continuation to coverage recovers EM.} On the HotpotQA
under-covered subset ($n{=}511$), EM roughly doubles from native-stop
$0.162$ to $0.323$ when forced to continue to \all coverage ($+16.0$\,pp),
and $57\%$ of the subset reaches \all when forced. Continuing to coverage
also beats continuing to the full budget ($0.311$ EM at $\sim\!8$ loops,
i.e.\ $+1.2$\,pp worse at more loops): \halt is not simply ``search
more'', and in this comparison the coverage point is the better
stopping target than the full budget.

Table~\ref{tab:continue} reports the full-population effect of
selectively continuing only the flagged under-covered subset (the rest
of the 7B population stays at its native stop): EM rises on all six
dataset$\times$claim-source cells with strictly positive paired
95\% CIs, and generated claims (no gold metadata) are nearly as strong
as gold, so the effect is not gold-dependent.

\begin{table}[t]
\centering
\small
\setlength{\tabcolsep}{2pt}
\begin{tabular}{lrr}
\toprule
Dataset & Full-pop $\Delta$EM (gold) & Full-pop $\Delta$EM (gen) \\
\midrule
HotpotQA & $+8.2\,$pp $[+6.2,+10.3]$ & $+7.1\,$pp $[+5.1,+9.2]$ \\
2Wiki    & $+2.7\,$pp $[+0.8,+4.5]$  & $+2.8\,$pp $[+0.9,+4.6]$ \\
MuSiQue  & $+4.3\,$pp $[+2.7,+5.9]$ & $+3.6\,$pp $[+2.2,+5.2]$ \\
\bottomrule
\end{tabular}
\caption{\textbf{Bidirectional control on the 7B agent.} Full-population
$\Delta$EM from selectively forcing continuation on \halt-flagged
under-covered trajectories only, keeping the rest of the population at
native stop. 95\% paired-bootstrap CIs (seed 13, 1000 iters); all six
cells are strictly positive.}
\label{tab:continue}
\end{table}

\paragraph{Mechanism control.} The gain tracks coverage recovery, not
just extra search: on HotpotQA, questions whose coverage recovered gain
$+26.2$\,pp EM versus $+2.7$\,pp for those that do not ($+7.7$ vs
$+0.9$\,pp on 2Wiki; $+15.6$ vs $+3.1$\,pp on MuSiQue) --- the gain
concentrates exactly where the coverage signal predicts it should
(subset sizes in Appendix~\ref{sec:appendix-continue}).

Together with \S\ref{sec:results-primary}--\ref{sec:results-7b}, this
shows the same coverage signal controls both directions of the stopping
decision: \emph{cut} on an over-searching 3B agent while preserving EM
under non-inferiority testing, \emph{extend} to coverage on an
under-searching 7B agent for a population EM gain.

\begin{table*}[t]
\centering
\footnotesize
\setlength{\tabcolsep}{4pt}
\begin{tabular}{lrrrrrl}
\toprule
& & \multicolumn{3}{c}{$\Delta$loops} & & \\
\cmidrule(lr){3-5}
Method & $n$ & 2Wiki & HotpotQA & MuSiQue & EM outcome & Frontier? \\
\midrule
\multicolumn{7}{l}{\emph{(A) Verifier-based stopping (ours)}} \\
\cmidrule(lr){1-7}
\rowcolor{halthl}
\quad\halt \all (gold)            & 1000 & $-37\%$ & $-45\%$ & $-20\%$ & NI (2/3 gold; 5/6 family) & \textbf{Yes} \\
\quad\halt \all (gen, comparator) & 1000 & $-17\%$ & $-18\%$ & $-2\%$  & NI (3/3 gen PASS) & Yes(gen) \\
\cmidrule(lr){1-7}
\multicolumn{7}{l}{\emph{(B) Trained adaptive-stopping baselines (gen-claim)}} \\
\cmidrule(lr){1-7}
\quad V-StopRAG (combined-3 train) & 1000 & ${\sim}0\%$ & ${\sim}0\%$ & ${\sim}0\%$ & tie & No$^\dagger$ \\
\quad V-SIMRAG                     & 1000 & ${\sim}0\%$ & ${\sim}0\%$ & ${\sim}0\%$ & tie & No$^\dagger$ \\
\quad V-StopRAG (2Wiki-only train) &  500 & $-21\%$     & (n/a)       & (n/a)       & $-1.2\,$pp & No$^\ddagger$ \\
\quad V-LLM-judge (zero-shot)      & 1000 & $-13\%$     & $-28\%$     & $-10\%$     & \halt NI vs judge 3/3 & No$^\S$ \\
\cmidrule(lr){1-7}
\multicolumn{7}{l}{\emph{(C) Adaptive-retrieval baselines}} \\
\cmidrule(lr){1-7}
\quad FLARE ($\beta{=}0.4$)          & 1000 & $-76\%$ & $-79\%$ & $-72\%$ & $-0.1/{-}6.3/{-}3.2\,$pp & No$^\P$ \\
\quad DRAGIN-approx ($\beta{=}0.6$)  & 1000 & $-43\%$ & $-50\%$ & $-38\%$ & NI (0/3 FAIL) & No$^\star$ \\
\cmidrule(lr){1-7}
\multicolumn{7}{l}{\emph{(D) Ensemble}} \\
\cmidrule(lr){1-7}
\quad \halt OR V-StopRAG & 1000 & $-17\%$ & $-18\%$ & $-2\%$ & $+0.3/{-}0.1/{-}0.1\,$pp & $\equiv$\,\halt$^\|$ \\
\bottomrule
\end{tabular}
\caption{\textbf{\halt vs adaptive controls on the
EM-non-inferiority $\times$ loop-superiority frontier.}
Groups~(A), (B), (D): identical Self-Ask 3B trajectories,
generated-claim setting (gold-claim \halt row is diagnostic);
group~(C) runs its own retrieval loops.
Threshold sweeps in
Figure~\ref{fig:baseline-threshold-sweep-all3}.
$^\dagger$: no loop savings;
$^\ddagger$: EM cost beyond $\varepsilon{=}2\,$pp;
$^\star$: $\Delta$EM within $\varepsilon$ but paired NI fails $0/3$
after Holm;
$^\S$: fires prematurely on HotpotQA, where \halt is F1-superior;
$^\P$: NI established on no dataset;
$^\|$: passes NI $3/3$, but is not a distinct operating point --- it
inherits \halt (gen)'s firings.}
\label{tab:baselines}
\end{table*}

\subsection{EM--loops Pareto and operating-point comparison}
\label{sec:results-novelty}

\begin{figure*}[t]
\centering
\includegraphics[width=0.78\textwidth]{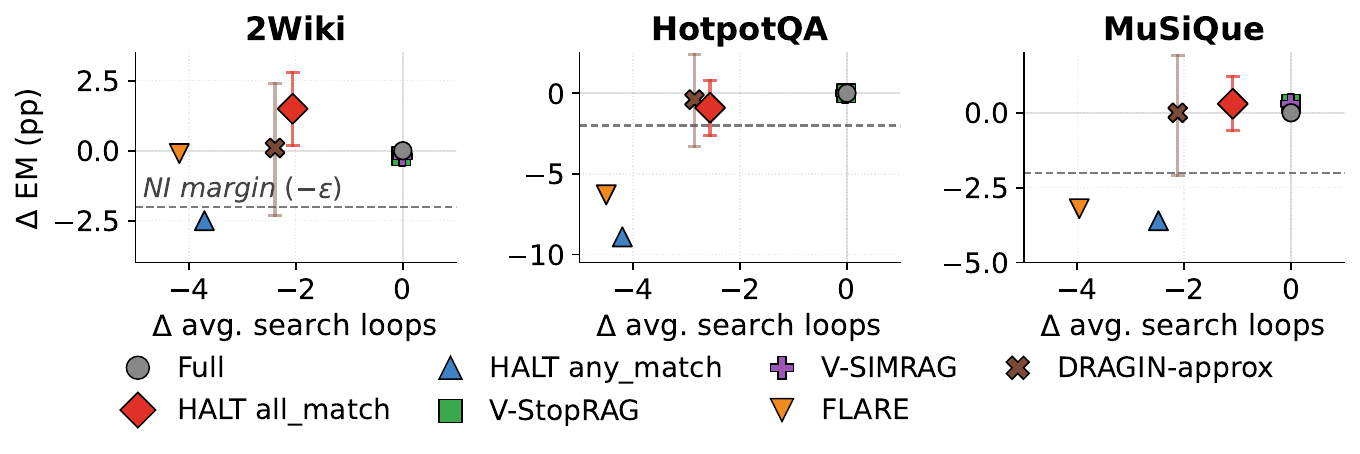}
\caption{\textbf{EM--search-loop tradeoff relative to \full,
3B Self-Ask agent.}
Each point shows the change in average search loops and Std-Ext EM
against \full on $1{,}000$ examples per dataset (\halt \all on MuSiQue
uses its $885$ valid rows). Lower $x$ means fewer loops, higher $y$
higher EM. The dashed line marks the non-inferiority margin
$-\varepsilon{=}-2\,$pp. Vertical bars show 95\%
paired-bootstrap CIs for $\Delta$EM on \halt \all and DRAGIN-approx.}
\label{fig:pareto}
\end{figure*}

Table~\ref{tab:baselines} compares \halt with adaptive-stopping and
adaptive-retrieval controls on identical Self-Ask 3B trajectories, and
Figure~\ref{fig:pareto} shows the corresponding EM--loops tradeoff.

With generated claims, \halt reduces loops while passing paired EM
non-inferiority on all three datasets. The gold-claim diagnostic row
shows the larger savings available when hop targets are clean. The
trained adaptive-stopping baselines either rarely fire on the combined
dataset distribution or lose EM when specialized to a single dataset.
FLARE reduces retrieval aggressively but establishes EM
non-inferiority on no dataset: it loses $6.3$/$3.2\,$pp of EM on
HotpotQA/MuSiQue, and its 2Wiki tie ($-0.1\,$pp) has a paired CI that
still crosses $-\varepsilon$.

\paragraph{Zero-shot sufficiency judge.}
Prompting the same frozen 3B backbone as a zero-shot sufficiency judge
(stop at the first loop with $P(\text{Yes})>0.95$) tests whether the
trained verifier is necessary. \halt is EM non-inferior to the judge on
all three datasets (paired, every 95\% CI lower bound above
$-\varepsilon$) and F1-superior on HotpotQA ($\Delta$F1 $+1.9$\pp, CI
$[+0.6,+3.2]$), where the judge fires on 43\% of questions but
prematurely: per-hop coverage is at least as good everywhere and better
where a global sufficiency prompt over-fires.

DRAGIN-approx has small point-estimate EM changes, but it does not pass
the same paired-bootstrap NI test used for \halt on any dataset after
Holm correction. For example, on HotpotQA its $\Delta$EM 95\% CI is
$[-3.3,+2.4]\,$pp. Under our implemented control suite, the standardized-extractor
endpoint, and identical Self-Ask 3B trajectories, \halt is the only
\emph{distinct} stopping policy that both reduces search loops and
passes the formal EM non-inferiority test.

The OR-ensemble does not improve the tradeoff. Because V-StopRAG
contributes almost no additional stops, its stop loop differs from
\halt (gen) on only $7/6/4$ of $1000$ questions; it therefore passes EM
non-inferiority ($+0.3/{-}0.1/{-}0.1\,$pp) by inheriting \halt's firings
rather than by occupying a separate point on the Pareto frontier.

\paragraph{Operating-point sweep.}
To check whether V-StopRAG and V-SIMRAG are disadvantaged by a single
threshold, we sweep both over
$\tau \in \{0.05, 0.10, 0.15, 0.20, 0.30, 0.50\}$ on the same
trajectories. No threshold preserves EM on all three datasets while
giving meaningful loop savings, and thresholds that help one dataset
hurt another: $\tau{=}0.10$ improves the 2Wiki tradeoff but pushes
HotpotQA beyond the $\varepsilon$ margin. The full sweep is in
Figure~\ref{fig:baseline-threshold-sweep-all3}.

\paragraph{Per-question divergence from V-SIMRAG.}
The stop loop chosen by \halt and V-SIMRAG differs on 56.6\%, 66.2\%,
and 27.7\% of questions on 2Wiki, HotpotQA, and MuSiQue, all above the
25\% pre-analysis divergence target: per-hop claim--evidence
verification is not simply recovering the same signal as a generic
sufficiency critic. Full divergence statistics, together with
fixed-budget, operating-point and verifier-logit diagnostics, are in
Appendices~\ref{sec:appendix-stats} and~\ref{sec:appendix-extended}.

\subsection{Extractor robustness}
\label{sec:results-extractor}

The main EM endpoint uses a standardized Qwen2.5-3B extractor so that
stopping policies share a fixed answer head. Rerunning the same
Self-Ask 3B trajectories with a Qwen2.5-7B-Instruct extractor
(Table~\ref{tab:stronger-extractor}) raises absolute EM by
$3$--$13\,$pp, but the \halt--\full gap stays small
($-0.5$/$-0.7$/$+0.4\,$pp on 2Wiki/HotpotQA/MuSiQue): the
EM-preservation result is not specific to the 3B extractor.
Point estimates are in \S\ref{sec:appendix-stats}.

\subsection{Retrieval-shift stress test: Setting~B}
\label{sec:results-settingB}

We also run an open-corpus pilot with dense E5 retrieval on Wikipedia
2018. On HotpotQA ($n{=}1000$, gold-claim), moving from the closed BM25
pool to open Wikipedia retrieval drops the agent's Std-Ext EM from
$0.306$ to $0.109$. \halt
responds conservatively: its \all fire rate falls from $67.6\%$ in
Setting~A to $6.4\%$ ($18.4\%$ for \any), with population
$\Delta\text{EM}{=}0\,$pp and $-4\%$ loops.

Rather than a failure mode, the fire rate is itself a signal: this is
verification-aware \emph{selective prediction}. On the $6.4\%$ of
questions where \all fires, same-extractor EM is $0.391$ versus the
$0.109$ population baseline: a paired-bootstrap $\Delta\text{EM}$ of
$+28.2\,$pp $[+16.3,+40.3]$, a $3.58\times$ $[2.55,4.78]$ lift. The
\any point still lifts EM by $+15.2\,$pp $[+9.4,+21.1]$
($2.39\times$) at $18.4\%$ coverage (seed 13, 1000 iters). \halt
identifies the answerable subset and abstains elsewhere. Bamboogle ($n{=}125$, generated
claims) behaves the same way: \all fires on $2.4\%$ of questions, with
EM unchanged on the fired subset and over the population
($0.072\to0.072$) --- highly selective, not a total abstention. Details
and a case study are in Appendix~\ref{sec:appendix-settingB-details}.

\begin{table}[t]
\centering
\small
\setlength{\tabcolsep}{4pt}
\begin{tabular}{lcccc}
\toprule
& \multicolumn{2}{c}{3B extractor} & \multicolumn{2}{c}{7B extractor} \\
\cmidrule(lr){2-3}\cmidrule(lr){4-5}
Dataset & Full & \halt & Full & \halt \\
\midrule
2Wiki    & 0.184 & 0.199 & 0.305 & 0.300 \\
HotpotQA & 0.306 & 0.297 & 0.431 & 0.424 \\
MuSiQue  & 0.083 & 0.086 & 0.120 & 0.124 \\
\bottomrule
\end{tabular}
\caption{\textbf{Extractor robustness.}
Same Self-Ask 3B trajectories scored with 3B and 7B standardized
extractors; \halt's gap to \full stays within $\pm0.7\,$pp everywhere.
MuSiQue is over its $885$ valid rows (\S\ref{sec:appendix-stats}).}
\label{tab:stronger-extractor}
\end{table}

\subsection{Cost and latency overhead}
\label{sec:cost}

\halt adds verifier calls while avoiding generator search loops. Raw
token accounting is net \emph{positive} for \halt on all six arms
(net $+4{,}993$ to $+8{,}503$ raw tokens/query): the verifier's cheap
parallel-prefill scoring processes more raw tokens than the expensive
sequential decode it avoids. We therefore state our cost claim in
search iterations and wall-clock, not raw tokens. Measured end-to-end
on HotpotQA ($n{=}100$, timed replay), \full averages $47.40\,$s/query
versus $30.94\,$s ($-35\%$) for \halt-gold and $40.62\,$s ($-14\%$)
for \halt-gen, with 95\% paired-bootstrap CIs strictly below zero in
both arms; the verifier is only $5$--$7\%$ of total time. Full tables
are in Appendix~\ref{sec:appendix-cost}.

\section{Analysis}
\label{sec:analysis}

We now test whether \halt stops for the intended reason: cumulative
evidence covers the expected hop claims. With the
fixed-budget comparison in \S\ref{sec:results-novelty}, the analyses
below check four simpler explanations: a fixed trajectory position, a
single matched hop, page-title overlap, and lexical overlap (further
diagnostics in Appendix~\ref{sec:appendix-extended}).

\paragraph{\all vs \any.}
The conservative \all policy produces a non-degenerate stopping
distribution: on Self-Ask~3B, \all firings spread over loops 1--3 with
a ``never'' tail, while the permissive \any fires at loop~1 on
$73/77/42\%$ of 2Wiki/HotpotQA/MuSiQue questions (each over its valid
rows) --- matching one hop is often easy on two-hop questions but
does not show that all required hops are covered, and MuSiQue's lower
rate reflects its longer claim chains. We default
to \all; per-loop distributions are in
Appendix~\ref{sec:appendix-extended}.

\subsection{Does \halt use claim--evidence alignment?}
\label{sec:verifier-ablations}

Three controlled ablations on the same Self-Ask~3B trajectories test
this directly (full table and protocols in
Appendix~\ref{sec:appendix-ablations}). Replacing a question's hop
claims with another question's collapses the \all fire rate
($0.591 \to 0.034$ on 2Wiki, $0.676 \to 0.196$ on HotpotQA); restoring
only the gold \textsc{Expected page} field recovers little
($0.034 \to 0.105$ on 2Wiki); and shuffling word order within retrieved
chunks also collapses firing ($0.676 \to 0.077$ on HotpotQA). None of
the four simpler explanations survives.

\paragraph{Verifier transfer.}
Trained only on HotpotQA discrepancy labels, the same checkpoint
reaches 0.940 macro-F1 on a 2Wiki discrepancy set built the same way
(vs.\ 0.930 on HotpotQA dev), and its page overlap with the HotpotQA
training pages is low on 2Wiki and MuSiQue --- transfer is not
explained by memorized page titles
(Appendix~\ref{sec:appendix-verifier}).
\section{Conclusion}

We presented \halt, a verification-aware stopping policy for
retrieval-augmented search agents. \halt leaves the host agent,
retriever, and generator unchanged, and stops when retrieved evidence
covers each expected reasoning hop. Across multi-hop QA benchmarks,
\halt reduces redundant search while largely preserving
standardized-extractor EM. Under open-corpus retrieval shift it
becomes cautious: on HotpotQA its fire rate drops from $67.6\%$ to
$6.4\%$, and on Bamboogle to $2.4\%$, both without EM loss.

The distinction is simple: stopping should ask not
only whether the generator appears ready to answer, but whether
retrieved evidence covers the question's reasoning structure. Making
this check per-hop reduces post-coverage search without retraining the
agent or relying on generator confidence.

\section*{Limitations}
\label{sec:limitations}

\paragraph{Scope of the stopping formulation.}
\halt assumes that the host agent exposes loop-level retrieval
trajectories and that expected reasoning hops can be represented as
claims. This fits structured agents such as Self-Ask, but does not directly
cover free-form agent trajectories without clear hop boundaries. Extending \halt to such settings would require either
runtime hop segmentation or a trajectory-level variant of the
claim--evidence alignment signal. The claim set is also fixed before
search begins; re-generating claims conditioned on the accumulated
evidence is a natural extension that we leave to future work.

\paragraph{Evaluation scope.}
Our main statistical claims are made in Setting~A, a controlled
closed-pool retrieval setting designed to isolate stopping from
open-corpus retrieval failure. Setting~B provides an open-corpus stress
test, where \halt behaves conservatively under retrieval shift, but we
do not treat it as the primary efficiency setting. Our open-corpus
selective-prediction analysis moreover uses gold hop claims as a
diagnostic; an open-corpus evaluation of the deployable generated-claim
condition remains future work. The reported results
should therefore be read as evidence that evidence-coverage stopping can
reduce redundant search in controlled multi-hop QA, not as a complete
deployment study across retrieval backends and domains.

\paragraph{Scale dependence of the savings.}
The efficiency gains depend on how much the host agent over-searches.
Our 3B Self-Ask agent rarely halts on its own---it runs to the search
budget on 78--88\% of questions---and \halt removes much of that
redundancy, whereas the 7B agent already stops natively on nearly all
questions, leaving little headroom, and \halt's savings shrink
accordingly. We do not read this as evidence that scale removes the need
for stopping control: even reward-trained search agents exhibit over-
and under-search \citep{wu2025hiprag,wu2025searchwisely}, and the small
7B effect may reflect our verifier under-capturing residual redundancy
rather than its absence. Characterizing that residual redundancy at
larger scales is future work.

\paragraph{Generality and deployment.}
The verifier is trained on HotpotQA discrepancy labels and reused across
datasets and agent scales. Although the transfer and ablation results
are encouraging, broader evaluation is needed across languages, domains,
agent formats, and claim-generation methods. In particular, all results
use 3B--7B open-weight backbones (with a cross-family transfer check on
Phi-3.5-mini); frontier-scale agents, whose search and stopping behavior
may differ qualitatively, remain untested. Our baseline comparisons
also include practical approximations for methods whose original
signals are not directly available in our setup. Finally, our cost
analysis reports measured end-to-end wall-clock with batched
verification on one dataset; a production deployment would still require
measurements under realistic scheduling and serving-stack constraints.
\section*{Ethics Statement}

\paragraph{Use of Scientific Artifacts}
Our research leveraged open-source tools including PyTorch
\citep{paszke2019pytorch} and the Huggingface ecosystem \citep{wolf2020transformers}
(\texttt{transformers}, \texttt{datasets}, \texttt{accelerate},
\texttt{peft}, \texttt{trl}), alongside pre-trained language models
from the Qwen2.5 family \citep{qwen2_5} obtained via the Huggingface
Hub. Retrieval is implemented with BM25 \citep{robertson2009bm25}
for Setting~A and the dense E5 encoder \citep{wang2022e5} over the
Wikipedia-2018 snapshot \citep{petroni2021kilt} for Setting~B. All
experiments use publicly available multi-hop QA benchmarks
(HotpotQA \citep{yang2018hotpotqa}, 2WikiMultihopQA
\citep{ho2020constructing}, MuSiQue \citep{trivedi2022musique}, and
Bamboogle \citep{press2023selfask}) under their respective licenses;
no human subjects were involved.

\paragraph{Use of AI Assistants}
We used AI coding assistants to help write experiment orchestration
and analysis scripts, and to assist with copy-editing the manuscript.
All experimental designs, claims, analyses, and final text were
authored and verified by the authors. No AI assistant was used to
generate experimental data, results, or conclusions.

\bibliography{references}

\newpage
\appendix
\section{Verifier Details}
\label{sec:appendix-verifier}

\paragraph{Discrepancy label construction.}
Each training example pairs an expected hop claim $c$ (built from a
HotpotQA supporting-fact title and its associated sentence) with a
candidate evidence sentence $e$ and receives one of three labels:
(i) \textsc{match} if $e$ is the gold supporting sentence for $c$ and
the supporting-fact title in $c$'s record matches the gold page;
(ii) \textsc{partial} if $e$ is on the same gold page as $c$ but
is \emph{not} the gold supporting sentence (topical overlap without
establishing the specific fact); and (iii) \textsc{null} if $e$ is
drawn from a benchmark distractor passage that does not appear in the
question's supporting facts. We negative-sample distractors uniformly
from the per-question distractor pool and balance the three classes
by undersampling \textsc{null} to match the smaller class count. The
train/dev split is question-disjoint at the HotpotQA question level,
and the same procedure (applied to 2Wiki supporting-fact metadata)
yields the cross-dataset evaluation set used in
Table~\ref{tab:cross-dataset-verifier}.

\paragraph{Architecture and training.}
The verifier is a Qwen2.5-3B-Instruct model fine-tuned via LoRA
\citep{hu2022lora} to emit one of the three label strings; at
inference each label is scored by its length-normalized token
log-probability and the argmax is predicted (no separate
classification head; \S\ref{sec:verifier}). We use rank
$r{=}16$, $\alpha{=}32$, dropout $0.05$, and target modules
$\{q,k,v,o,\text{gate},\text{up},\text{down}\}_{\text{proj}}$.
The training data is a question-disjoint balanced subset of HotpotQA
discrepancy labels ($9{,}626$ training rows / $2{,}374$ dev rows over
\textsc{match}/\textsc{partial}/\textsc{null}). We train for
$1{,}000$ steps with paged AdamW (8-bit), learning rate
$2{\times}10^{-4}$, per-device batch size $1$, and $4$-bit NF4
quantization with double quantization and bf16 compute. The held-out
dev macro-F1 is $0.9304$.

We run input-component ablations on the held-out HotpotQA dev split.
Masking the answer surface drops macro-F1 by only $0.002$, suggesting
that the verifier is not relying on answer-string matching. Dropping
the \textsc{Expected page} field drops macro-F1 by $0.176$; dropping
the \textsc{Expected hop claim} field drops macro-F1 by $0.218$; and
lowercasing only the evidence sentence drops macro-F1 by $0.009$.
The expected claim is therefore the most load-bearing input component,
with page information contributing additional signal.

\paragraph{Cross-dataset verifier macro-F1.}
The verifier is trained on HotpotQA discrepancy labels only. To test
label-level transfer, we evaluate the same checkpoint on a 2Wiki
discrepancy-label set constructed with the same procedure
($7{,}003$ rows over the same three labels). MuSiQue does not ship
supporting-fact metadata in our gold file, so we cannot build a directly
comparable label set for it.

\begin{table}[!ht]
\centering
\small
\setlength{\tabcolsep}{4pt}
\begin{tabular}{lrr}
\toprule
Eval split & Acc.\ & Macro-F1 \\
\midrule
HotpotQA dev ($n{=}2374$, in-dom.)  & 0.946 & 0.930 \\
2Wiki dev ($n{=}7003$, cross-dset)  & 0.942 & 0.940 \\
\bottomrule
\end{tabular}
\caption{Cross-dataset verifier classification quality. The
HotpotQA-trained verifier transfers to the 2Wiki label set at
macro-F1 $0.940$, essentially matching the in-domain $0.930$.}
\label{tab:cross-dataset-verifier}
\end{table}

\paragraph{Page-overlap diagnostic.}
A possible concern is that the strong 2Wiki transfer could be explained
by Wikipedia-page memorization rather than claim--evidence matching.
We therefore measure page overlap between HotpotQA verifier-training
pages and evaluation supporting-fact pages. From the HotpotQA SFT data
($\approx 12\text{K}$ examples), we extract the set of unique
``Expected page'' / ``Evidence page'' titles ($n{=}12{,}070$) and
compute overlap for each evaluation set.

\begin{table}[!ht]
\centering
\scriptsize
\setlength{\tabcolsep}{3pt}
\begin{tabular}{lrrr}
\toprule
Eval set & Page overlap & Any-pg Q & All-pg Q \\
\midrule
HotpotQA $(n{=}1000)$ & $1198/1976$ ($60.6\%$) & $85.1\%$ & $36.8\%$ \\
2Wiki    $(n{=}1000)$ & $53/2343$ ($2.3\%$)    & $5.6\%$  & $0.4\%$ \\
MuSiQue  $(n{=}911)$  & $10/950$ ($1.1\%$)     & $3.3\%$  & $0.8\%$ \\
\bottomrule
\end{tabular}
\caption{Page-level overlap between HotpotQA verifier-training pages
and evaluation supporting-fact pages. Cross-dataset evaluation pages
are almost disjoint from the verifier-training pages, especially under
the all-page question-level criterion. The $n$ in each row label is
the number of evaluation questions with extractable supporting-fact
page metadata (89/1000 MuSiQue questions lack this metadata and are
excluded); the ``Page overlap'' fraction's denominator is instead the
number of \emph{unique} evaluation pages pooled across those questions
(950 for MuSiQue), not a per-question count, so it need not match $n$.}
\label{tab:page-overlap}
\end{table}

\begin{figure*}[!t]
\centering
\includegraphics[width=\textwidth]{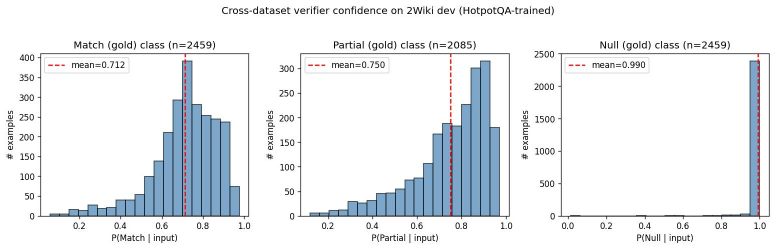}
\caption{Cross-dataset verifier confidence on 2Wiki dev
($n{=}7003$). Histograms show $P(\text{gold label})$ for each class;
y-scales are independent per panel.}
\label{fig:verifier-conf}
\end{figure*}

\paragraph{Bidirectional cross-training (2Wiki$\to$HotpotQA).}
The HotpotQA-trained verifier obtains macro-F1 $0.940$ on 2Wiki,
slightly above its in-domain $0.930$. To check whether this asymmetry
comes from an easier 2Wiki label distribution, we train the same LoRA
verifier from scratch on 2Wiki SFT data and evaluate both in-domain
and cross-domain. The 2Wiki-trained run uses $100$ training steps, with
otherwise identical hyperparameters.

\begin{table}[!ht]
\centering
\footnotesize
\setlength{\tabcolsep}{3pt}
\begin{tabular}{llrr}
\toprule
Train set & Eval set & Acc.\ & Macro-F1 \\
\midrule
HotpotQA SFT & HotpotQA (in-dom.) & 0.946 & 0.930 \\
HotpotQA SFT & 2Wiki (cross)      & 0.942 & 0.940 \\
2Wiki SFT    & 2Wiki (in-dom.)    & 0.914 & 0.911 \\
2Wiki SFT    & HotpotQA (cross)   & 0.860 & 0.859 \\
\bottomrule
\end{tabular}
\caption{Bidirectional cross-training of the LoRA verifier. The
2Wiki-trained verifier drops $5.2$~pp macro-F1 on HotpotQA, while
the reverse direction shows no drop.}
\label{tab:bidirectional-verifier}
\end{table}

\paragraph{Verifier confidence on cross-dataset transfer.}
Figure~\ref{fig:verifier-conf} shows the verifier's softmax
probability of the gold label on the 2Wiki cross-dataset evaluation.
The Null class is near-saturated (mean $0.99$); Match and Partial are
less certain (mean $0.71$ and $0.75$) and form the main error boundary.

\FloatBarrier

\section{Per-cell Statistical Test}
\label{sec:appendix-stats}

\paragraph{NI margin choice.}
The non-inferiority margin $\varepsilon=0.02$ (2\,pp) used in
\S\ref{sec:stats} is fixed in absolute EM points rather than relative
to each dataset's baseline EM. This keeps the decision rule identical
across datasets. It is also conservative on low-EM datasets: for
MuSiQue, where the baseline EM is around $0.07$, a 2\,pp margin is a
large relative change. We also pre-specify a 25\% per-question
stop-loop divergence target for the V-SIMRAG comparison; the
corresponding divergence summary appears below.

\paragraph{MuSiQue denominator.} In the gold-claim condition, $115$ of
the $1{,}000$ sampled MuSiQue questions yield no atomic hop claims
(\texttt{skip\_reason:~no\_claims}) and carry no verifier trajectory at
all. They are excluded from both numerator and denominator, so every
MuSiQue gold-claim point estimate and test in this paper is reported
over the remaining $n{=}885$ valid rows; the generated-claim condition
always produces at least one claim and stays at $n{=}1000$, and 2Wiki
and HotpotQA have no such rows in either condition. The excluded rows contribute exactly zero signal to the
comparison (EM $0.000$ and F1 $0.042$ under both \full and \halt, hence
$\Delta{=}0$ on every endpoint), so retaining them would move the level
columns by about $1$\,pp while leaving the $\Delta$ columns and every
verdict unchanged. Separately, the MuSiQue $n{=}1000$ evaluation slice
contains only $911$ unique question ids, because $89$ questions were
sampled and scored twice; paired statistics therefore pair rows
positionally rather than by question id, which is exact because every
MuSiQue artifact shares the same row order. Finally, the valid-row set
is determined per run by whether claim generation returns claims for a
question, so its exact membership can differ slightly between runs: the
original 7B gold-claim run yields $105$ distinct no-claims questions
and its forced-continuation counterpart $106$, and this is not a single
flip --- $11$ questions are skips only in the forced run and $10$ only
in the original. We therefore treat the valid-row set as a per-run
property of the artifact being analysed rather than as a fixed subset
of MuSiQue.

\subsection{3B Family (Holm $m{=}12$)}

\begin{table}[!ht]
\centering
\scriptsize
\setlength{\tabcolsep}{2pt}
\resizebox{\columnwidth}{!}{%
\begin{tabular}{lrrcrcc}
\toprule
Cell & $\Delta$EM & 95\% CI EM & $p_\text{NI}$ & $\Delta$loops & $p_\text{sup}$ & Pass \\
\midrule
2Wiki gen     & $+0.003$ & $[-0.007, +0.013]$ & 0.000 & $-0.953$ & 0.000 & PASS \\
2Wiki gold    & $+0.015$ & $[+0.002, +0.028]$ & 0.000 & $-2.053$ & 0.000 & PASS \\
HotpotQA gen  & $-0.001$ & $[-0.013, +0.011]$ & 0.001 & $-1.012$ & 0.000 & PASS \\
HotpotQA gold & $-0.009$ & $[-0.026, +0.008]$ & 0.095 & $-2.557$ & 0.000 & FAIL \\
MuSiQue gen   & $-0.001$ & $[-0.003, +0.000]$ & 0.000 & $-0.115$ & 0.000 & PASS \\
MuSiQue gold  & $+0.003$ & $[-0.006, +0.012]$ & 0.000 & $-1.089$ & 0.000 & PASS \\
\bottomrule
\end{tabular}}
\caption{3B family non-inferiority + superiority test,
$\varepsilon{=}2\,$pp, Holm $m{=}12$. PASS requires rejection on both
endpoints. The single failing cell is HotpotQA gold, whose EM CI
crosses $-\varepsilon$.}
\label{tab:ni-3b}
\end{table}

\paragraph{Paired-binary robustness.}
Since EM is binary at the question level, we additionally compute
Newcombe hybrid-score CIs for the paired proportion difference and
exact two-sided McNemar tests on discordant pairs. These tests use the
same per-question EM outcomes as Table~\ref{tab:ni-3b}. The Newcombe
NI decisions exactly match the paired-bootstrap NI decisions; the only
EM-NI failure remains HotpotQA gold. McNemar is an equality test rather
than a non-inferiority test, so we report it as a direction-of-change
diagnostic rather than as the decision rule.

\begin{table}[!ht]
\centering
\scriptsize
\setlength{\tabcolsep}{2pt}
\resizebox{\columnwidth}{!}{%
\begin{tabular}{lrrrrcr}
\toprule
Cell & Full-only & \halt-only & $\Delta$EM & Boot CI & Newcombe CI & McNemar $p$ \\
\midrule
2Wiki gen     & 11 & 14 & $+0.3$ & $[-0.7,+1.3]$ & $[-0.7,+1.3]$ & 0.690 \\
2Wiki gold    & 15 & 30 & $+1.5$ & $[+0.2,+2.8]$ & $[+0.2,+2.8]$ & 0.036 \\
HotpotQA gen  & 17 & 16 & $-0.1$ & $[-1.3,+1.1]$ & $[-1.2,+1.0]$ & 1.000 \\
HotpotQA gold & 40 & 31 & $-0.9$ & $[-2.6,+0.8]$ & $[-2.6,+0.8]$ & 0.342 \\
MuSiQue gen   &  1 &  0 & $-0.1$ & $[-0.3,+0.0]$ & $[-0.5,+0.3]$ & 1.000 \\
MuSiQue gold  &  6 &  9 & $+0.3$ & $[-0.6,+1.2]$ & $[-0.6,+1.3]$ & 0.607 \\
\bottomrule
\end{tabular}}
\caption{Paired-binary robustness checks for 3B EM. Full-only and
\halt-only are discordant question counts. CIs are in percentage
points for $\Delta\text{EM}=\text{EM}_{\halt}-\text{EM}_{\full}$.
Newcombe non-inferiority at $\varepsilon{=}2$~pp gives the same
PASS/FAIL pattern as the bootstrap CI in Table~\ref{tab:ni-3b}.}
\label{tab:paired-binary-robustness}
\end{table}

\subsection{7B Family (Holm $m{=}6$)}

Table~\ref{tab:ni-7b} repeats the test for the 7B agent, where the
family is the three gold-claim cells on both endpoints.

\begin{table}[!ht]
\centering
\small
\setlength{\tabcolsep}{4pt}
\begin{tabular}{lrrcc}
\toprule
Cell & $n$ & $\Delta$EM & Rej.\ NI? & Rej.\ loops? \\
\midrule
2Wiki 7B    & 1000 & $+0.006$ & $\checkmark$ & $\checkmark$ \\
HotpotQA 7B & 1000 & $-0.005$ & $\checkmark$ & $\checkmark$ \\
MuSiQue 7B  & 1000 & $-0.002$ & $\checkmark$ & $\checkmark$ \\
\bottomrule
\end{tabular}
\caption{7B family non-inferiority + superiority test,
$\varepsilon{=}2\,$pp, Holm $m{=}6$. All three cells pass both
endpoints.}
\label{tab:ni-7b}
\end{table}

\subsection{7B-Extractor Family}
We repeat the standardized-extractor evaluation with
Qwen2.5-7B-Instruct on the same Self-Ask 3B trajectories. The point
estimates are the same as Table~\ref{tab:stronger-extractor}; we report
them as a robustness check rather than as a separate formal NI family.

\begin{table}[!ht]
\centering
\small
\setlength{\tabcolsep}{4pt}
\begin{tabular}{lrrr}
\toprule
Cell & $n$ & $\Delta$EM & $\Delta$ loops \\
\midrule
2Wiki 7B-ext    & 1000 & $-0.005$ & $-2.053$ \\
HotpotQA 7B-ext & 1000 & $-0.007$ & $-2.557$ \\
MuSiQue 7B-ext  &  885 & $+0.004$ & $-1.089$ \\
\bottomrule
\end{tabular}
\caption{Point-estimate changes under the stronger standardized
extractor used in Table~\ref{tab:stronger-extractor}. All EM changes
are within $\pm 0.7\,$pp.}
\label{tab:ni-7b-extractor}
\end{table}

\subsection{DRAGIN-Approx (Holm $m{=}3$)}
The DRAGIN-approx point-estimate $\Delta$EM falls within
$\varepsilon{=}2\,$pp on all three datasets
(\S\ref{sec:results-novelty}). We apply the same paired-bootstrap NI
test under Holm $m{=}3$ to check whether these point estimates imply
formal non-inferiority. They do not: all three 95\% CIs cross
$-\varepsilon$.

\begin{table}[!ht]
\centering
\scriptsize
\setlength{\tabcolsep}{2pt}
\begin{tabular}{lrrcrcc}
\toprule
Cell & $\Delta$EM & 95\% CI EM & $p_\text{NI}$ & $\Delta$loops & $p_\text{sup}$ & Pass \\
\midrule
2Wiki    & $+0.001$ & $[-0.023, +0.024]$ & 0.101 & $-2.388$ & 0.000 & FAIL \\
HotpotQA & $-0.004$ & $[-0.033, +0.024]$ & 0.161 & $-2.841$ & 0.000 & FAIL \\
MuSiQue  & $-0.001$ & $[-0.021, +0.019]$ & 0.078 & $-2.144$ & 0.000 & FAIL \\
\bottomrule
\end{tabular}
\caption{DRAGIN-approx vs \full paired NI + superiority,
$\varepsilon{=}2\,$pp, Holm $m{=}3$. All three datasets pass loop
superiority but fail EM non-inferiority because the 95\% CI for
$\Delta$EM crosses $-\varepsilon$.}
\label{tab:ni-dragin}
\end{table}

\subsection{Stop-Loop Divergence from V-SIMRAG}

Table~\ref{tab:diverge} reports the per-question stop-loop
divergence against the $25\%$ pre-analysis target quoted in
\S\ref{sec:results-novelty}.

\begin{table}[!ht]
\centering
\footnotesize
\setlength{\tabcolsep}{4pt}
\begin{tabular}{lrr}
\toprule
Dataset & Divergence & Target \\
\midrule
2Wiki     & 56.6\% & 25\% \\
HotpotQA  & 66.2\% & 25\% \\
MuSiQue   & 27.7\% & 25\% \\
\bottomrule
\end{tabular}
\caption{Per-question stop-loop divergence between \halt \all and
V-SIMRAG. All datasets exceed the pre-analysis target, supporting the
claim that per-hop claim--evidence verification is distinct from the
generic sufficiency critic.}
\label{tab:diverge}
\end{table}

\FloatBarrier

\section{Hop Claim Construction Details}
\label{sec:appendix-claim-quality}

\paragraph{Claim record schema.}
Gold and generated claims use the same record schema; only the source
of the fields differs. Each record contains a supporting page
\texttt{title} (gold) or LLM-generated \texttt{target\_entity}, an
\texttt{expected\_target} string, a \texttt{role} field
(\texttt{bridge\_title} vs.\ \texttt{answer\_node}), an
\texttt{answer\_type} field, and an \texttt{expected\_claim}. In the
gold condition, $K$ equals the number of distinct supporting titles.
In the generated condition, $K$ equals the length of the parsed JSON
list. The verifier sees the same input schema in both cases: question,
\texttt{expected\_target} (the hop claim's predicted target entity,
\emph{not} the question's gold answer), expected hop claim, expected
page/target fields, evidence page, and evidence sentence.

\paragraph{No gold-answer leakage at test time.}
The verifier's \texttt{expected\_target} field is the per-hop target
entity carried in the claim record, not the question's gold answer.
In the gold-claim condition this string is derived from
supporting-fact metadata for the corresponding hop (e.g.\ the bridge
title); in the generated-claim condition it is the
\texttt{expected\_target} produced by the LLM-generated claim list
(\S\ref{sec:appendix-claim-quality}). The question-level gold answer
is never an input. Consistent with this, masking the
\texttt{expected\_target} string at evaluation time drops verifier
macro-F1 by only $0.002$ (\S\ref{sec:appendix-verifier}), confirming
the verifier does not rely on answer-string matching even when the
field contains a label-bearing entity.

\paragraph{Generation prompt.}
Generated claims are produced by zero-shot prompting
Qwen2.5-3B-Instruct with the question and the instruction:
\begin{quote}\small
For the multi-hop question below, list the entities or facts that
must be retrieved \emph{in order} to answer it. Each item should
identify a target entity, what role it plays in the question
(bridge title, intermediate entity, or final answer node), and what
the retrieved evidence should establish about it. Output a JSON
list with fields \texttt{target\_entity}, \texttt{expected\_target},
\texttt{target\_role}, \texttt{answer\_type}, \texttt{expected\_claim}.
\end{quote}
We do not provide in-context examples. If JSON parsing fails, we fall
back to a single hop claim constructed from the question and the LLM's
first sentence.

\paragraph{Number of claims $K$.}
On 2Wiki generated claims, $K{=}2$ for $986$ questions, $K{=}3$ for
$10$, $K{=}4$ for $3$, and $K{=}5$ for $1$. On HotpotQA generated
claims, $K{=}1$ for $1$, $K{=}2$ for $992$, $K{=}3$ for $6$, and
$K{=}4$ for $1$. The two-hop majority reflects the structure of both
benchmarks.

\paragraph{Quality vs gold.}
Treating each question's gold supporting-fact titles as target
entities, generated-claim target-entity overlap yields
$F_1{=}0.701$ on 2Wiki and $F_1{=}0.455$ on HotpotQA. We also track a
\textsc{no\_match} bucket where none of the generated target entities
overlaps any gold title. \halt's fire rate in this bucket is $0\%$ on
both datasets, indicating that poor generated claims reduce firing
rather than inducing spurious stops.

\paragraph{Generation cost.}
Each question's generated claims require one $\leq 256$-token LLM call
before the search loop starts. In our implementation this adds
approximately $0.4\,$s per question and is included in the cost profile
(\S\ref{sec:cost}).

\FloatBarrier

\section{Reproducibility Details}
\label{sec:appendix-repro}

\paragraph{Hardware and software.}
All experiments run under Python~3.12 with PyTorch~2.7,
Transformers~4.53, sentence-transformers~5.0, and
\texttt{datasets}~4.0 on a workstation with two RTX~3090 (24~GB) and
one RTX~4090 (24~GB) GPU. The 3B agent and verifier each fit on a
single 24~GB GPU; the 7B agent is sharded across two 24~GB GPUs. The
Setting~B FAISS index is memory-mapped from host RAM, making dense
retrieval CPU-RAM-bound rather than GPU-bound. All experiments use
random seed~$13$.

\subsection{Baseline Reproduction}
\label{sec:appendix-baselines}

All baseline controls in Table~\ref{tab:baselines} operate on the same
Self-Ask 3B trajectories as \halt whenever applicable, so loop counts
and EM are directly comparable.

\paragraph{V-StopRAG.}
V-StopRAG is implemented as a Qwen2.5-1.5B-Instruct backbone with a
2-way classification head over trajectory states. A state is labeled
positive if adding more loops does not improve EM on the full
trajectory. We train two configurations: a combined-3 model pooled
across HotpotQA, 2Wiki, and MuSiQue, and a 2Wiki-only specialist.
Both use 2 epochs, learning rate $10^{-4}$, paged AdamW, 4-bit NF4
quantization, max input length $2048$, and inference threshold
$\tau{=}0.5$. The model fires if
$P(\text{stop}\mid\text{state})\geq\tau$ at any loop.

\paragraph{V-SIMRAG.}
V-SIMRAG uses the same backbone and training recipe as V-StopRAG, but
its input additionally includes the current sub-question and the
agent's loop-level intermediate answer. The positive class is whether
the current question, cumulative evidence, and intermediate answer are
sufficient to answer the original question.

\paragraph{FLARE.}
We reproduce the public FLARE control: the generator emits a sentence,
and if any token in that sentence has probability below
$\beta{=}0.4$, retrieval is triggered using the partial sentence as the
query. We use Qwen2.5-3B-Instruct, top-$K{=}3$ sentences per retrieval,
a maximum of 8 generated sentences per question, 64 new tokens per
sentence, 4-bit NF4 quantization, and seed $13$.

\paragraph{FLARE $\beta$-sweep robustness ($n{=}1000$).}
Because $\beta$ controls how aggressively FLARE retrieves, the
comparison in Table~\ref{tab:baselines} could be sensitive to that
single choice. We therefore re-run FLARE at three confidence thresholds
$\beta\in\{0.3, 0.4, 0.5\}$ on all three datasets and report
standardized-extractor EM in Table~\ref{tab:flare-beta-sweep}. EM
increases monotonically with $\beta$ on every dataset, confirming that
FLARE trades search depth for accuracy. Even at the most retrieval-heavy
setting ($\beta{=}0.5$), FLARE remains below the Self-Ask Full
reference on HotpotQA ($-3.8$~pp) and MuSiQue ($-0.6$~pp). On 2Wiki the
two are statistically tied ($+0.3$~pp), and at this setting FLARE
issues \emph{fewer} retrieval loops than \halt on that dataset
($2.02$ vs $4.56$ in the generated-claim condition). FLARE's cost is
therefore not search depth but the interface it requires and the
accuracy it gives up: it monitors per-token generation confidence and
regenerates the current sentence whenever retrieval is triggered, which
presumes white-box access to the generator's token probabilities, and
its EM is dataset-fragile --- as the rows above show, no single $\beta$
preserves EM on all three datasets. We therefore keep
FLARE as an adaptive-retrieval control rather than as a substitute for
the verifier stopping signal.

\begin{table}[!ht]
\centering
\small
\setlength{\tabcolsep}{4pt}
\begin{tabular}{lrrrr}
\toprule
$\beta$ & 2Wiki & HotpotQA & MuSiQue & Avg.\ \\
\midrule
$0.3$        & 0.179 & 0.210 & 0.037 & 0.142 \\
$0.4$ (main) & 0.181 & 0.243 & 0.044 & 0.156 \\
$0.5$        & 0.185 & 0.268 & 0.070 & 0.174 \\
\midrule
Self-Ask Full ref.\ & 0.182 & 0.306 & 0.076 & 0.188 \\
\bottomrule
\end{tabular}
\caption{FLARE standardized-extractor EM at $n{=}1000$ for
$\beta\in\{0.3, 0.4, 0.5\}$. EM rises with $\beta$ on every dataset, but
FLARE remains below Self-Ask Full on HotpotQA and MuSiQue at all three
thresholds.}
\label{tab:flare-beta-sweep}
\end{table}

\paragraph{DRAGIN approximation.}
DRAGIN's original RIND signal depends on token-level attention
statistics \citep{su2024dragin}. Because those hooks are not available
in our setup, DRAGIN-approx uses the FLARE interface with a higher
retrieval threshold, $\beta{=}0.6$. We treat it as an adaptive-retrieval
reference rather than as an exact reproduction of DRAGIN.

\paragraph{\halt OR V-StopRAG.}
The OR ensemble stops at the first loop where either \halt \all or
V-StopRAG fires. It is included only as a redundancy check for the stop
signals.

\subsection{Setting~B Retrieval Details}
\label{sec:appendix-settingB}

Setting~B replaces Setting~A's per-question BM25 distractor pool with
dense retrieval over the full Wikipedia 2018 corpus. The agent policy,
generator, and verifier are otherwise unchanged.

\begin{itemize}
\item \textbf{Corpus.} FlashRAG Wikipedia 2018 snapshot
  \citep{petroni2021kilt} ($\sim 7.9$M passages after segmentation).
\item \textbf{Query encoder.} \texttt{intfloat/e5-base-v2}
  \citep{wang2022e5} with the standard ``query:'' prefix.
\item \textbf{Index.} FAISS \texttt{IndexFlat} exact inner-product
  search, memory-mapped from the prebuilt \texttt{wiki18\_e5.index}.
\item \textbf{Retrieval.} Top-$K{=}3$ passages per follow-up query,
  with up to 12 sentences per passage after sentence segmentation.
\item \textbf{Agent.} Self-Ask Qwen2.5-3B-Instruct, 4-bit NF4
  quantization, max 8 loops, max 80 new tokens per follow-up, seed
  $13$.
\item \textbf{Devices.} QA generator on \texttt{cuda:0}, E5 encoder on
  \texttt{cuda:1}; FAISS search is CPU-RAM-bound.
\end{itemize}

Bamboogle uses the same retrieval stack and LLM-generated hop claims,
because it does not provide gold supporting-fact metadata.

\subsection{Standardized Extractor}
\label{sec:appendix-extractor}

To compare retrieval-side stopping policies under a common answer head,
we evaluate every stopping policy with a frozen standardized extractor
(Std-Ext) that takes the cumulative retrieved chunks at the stop loop
and emits a short-form answer. The 3B Std-Ext is Qwen2.5-3B-Instruct;
the stronger extractor used in \S\ref{sec:results-extractor} is
Qwen2.5-7B-Instruct. Both run with 4-bit NF4 quantization.

\paragraph{Prompt template.}
A two-turn chat with the system instruction
\textit{``You are a concise QA assistant. Output only the answer, no
extra words.''} The user turn is:
\begin{quote}\small\ttfamily
You will read evidence and answer a question concisely.\\[2pt]
Evidence:\\
\textit{<concatenated retrieved chunk text>}\\[2pt]
Question: \textit{<original question>}\\
Answer (concise, no explanation):
\end{quote}

\paragraph{Decoding.}
We use greedy decoding, $48$ new tokens, and max input length
$4{,}000$ tokens. If the chat-templated prompt exceeds the input
budget, we truncate only the evidence block; the system prompt,
question, and answer cue are preserved.

\paragraph{Why standardize.}
Without a fixed extractor, EM-vs-loops comparisons are confounded by
each agent's native answer head. On a 2Wiki $n{=}200$ diagnostic,
FLARE and DRAGIN-approx have near-zero native EM but Std-Ext EM of
$0.135$ and $0.145$, respectively. The standardized extractor therefore
compares stopping policies by the answerability of their retrieved
chunks rather than by implementation-specific answer formatting.

\FloatBarrier

\section{Extended Results and Robustness}
\label{sec:appendix-extended}

This appendix collects auxiliary analyses that support the Results and
Analysis sections.

\subsection{Fixed-budget baseline}
\label{sec:appendix-fixed-budget}

A simple alternative to learned stopping is to truncate every
trajectory at a fixed loop budget. We compare fixed budgets against
\halt on the same Self-Ask~3B trajectories in the generated-claim
condition.

\begin{table}[!ht]
\centering
\footnotesize
\setlength{\tabcolsep}{3pt}
\begin{tabular}{llrrr}
\toprule
& Policy & 2Wiki & HotpotQA & MuSiQue \\
\midrule
\multirow{4}{*}{EM}
  & fix=3                 & 0.184 & 0.294 & 0.066 \\
  & fix=5                 & 0.187 & 0.309 & 0.076 \\
  & Full                  & 0.182 & 0.306 & 0.076 \\
  & \halt \all            & \textbf{0.185} & 0.305 & 0.075 \\
\midrule
\multirow{4}{*}{loops}
  & fix=3                 & 3.00 & 3.00 & 3.00 \\
  & fix=5                 & 5.00 & 5.00 & 5.00 \\
  & Full                  & 5.52 & 5.71 & 5.52 \\
  & \halt \all            & \textbf{4.56} & \textbf{4.69} & 5.40 \\
\bottomrule
\end{tabular}
\caption{Fixed-budget baseline on Self-Ask~3B, generated-claim
condition. Fixed budgets occasionally match \halt but apply a uniform
cutoff and do not test evidence sufficiency per question.}
\label{tab:fixed-budget}
\end{table}

\subsection{Operating-point sensitivity}
\label{sec:appendix-operating-point}

\paragraph{Baseline threshold sweep.}
V-StopRAG and V-SIMRAG use a binary stop classifier with default
threshold $\tau{=}0.5$. On our trajectories this default rarely fires
(V-StopRAG $\leq 1\%$, V-SIMRAG $\leq 1.4\%$ across datasets). We
therefore re-threshold cached per-loop stop probabilities at
$\tau \in \{0.05, 0.10, 0.15, 0.20, 0.30, 0.50\}$.

\begin{table}[!ht]
\centering
\small
\setlength{\tabcolsep}{4pt}
\begin{tabular}{lrrrr}
\toprule
Method & $\tau$ & Fire & Avg.\ stop & EM \\
\midrule
\multicolumn{5}{l}{\emph{Baselines (2Wiki, Self-Ask~3B)}} \\
V-StopRAG & 0.05 & 0.98 & 1.76 & 0.161 \\
V-StopRAG & 0.10 & 0.80 & 3.02 & 0.188 \\
V-StopRAG & 0.20 & 0.32 & 4.75 & 0.186 \\
V-StopRAG & 0.50 & 0.01 & 5.49 & 0.182 \\
V-SIMRAG  & 0.05 & 0.98 & 1.58 & 0.158 \\
V-SIMRAG  & 0.10 & 0.83 & 2.98 & 0.175 \\
V-SIMRAG  & 0.20 & 0.31 & 4.82 & 0.182 \\
V-SIMRAG  & 0.50 & 0.00 & 5.51 & 0.182 \\
\midrule
\multicolumn{5}{l}{\emph{\halt (match-logit margin, same trajectories)}} \\
\halt $\tau{=}{-}2$ & --- & 0.895 & 1.85 & \textbf{0.183} \\
\halt $\tau{=}{-}1$ & --- & 0.758 & 2.65 & \textbf{0.196} \\
\halt $\tau{=}0$    & --- & 0.594 & 3.44 & \textbf{0.195} \\
\halt $\tau{=}1$    & --- & 0.149 & 5.03 & \textbf{0.186} \\
\halt $\tau{=}2$    & --- & 0.020 & 5.45 & 0.182 \\
\bottomrule
\end{tabular}
\caption{Operating-point sweep on 2Wiki. \halt stays on the
loop-vs-EM frontier over a broad range of match-logit thresholds,
whereas aggressive V-StopRAG/V-SIMRAG thresholds shorten trajectories
with larger EM loss.}
\label{tab:baseline-threshold-sweep}
\end{table}

\begin{figure*}[!t]
\centering
\includegraphics[width=\textwidth]{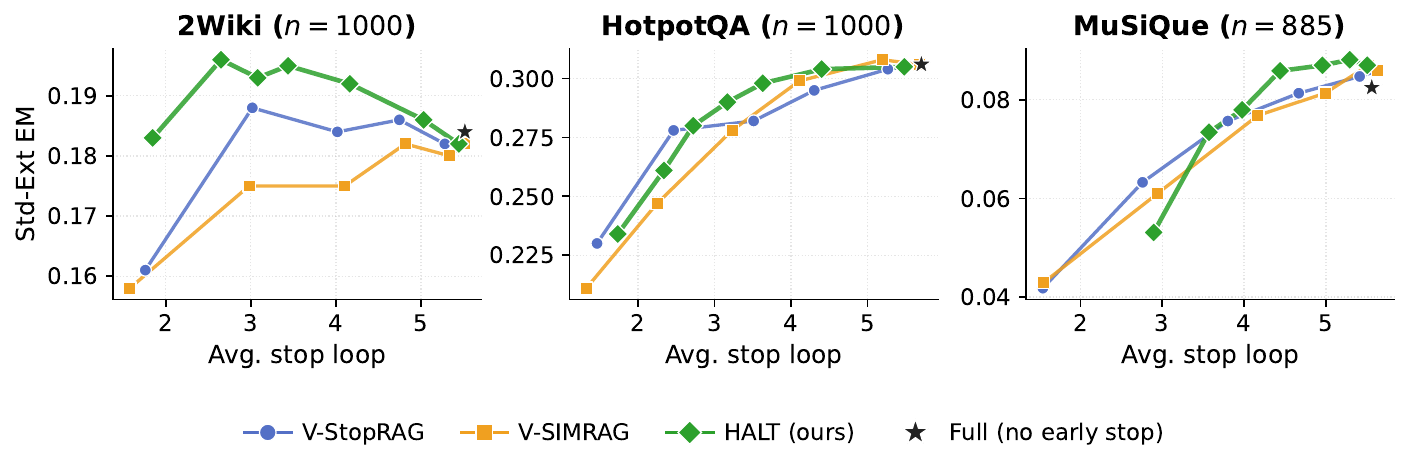}
\caption{\textbf{Operating-point sweep across all three datasets,
Self-Ask~3B.} V-StopRAG and V-SIMRAG are swept by post-hoc
re-thresholding cached stop probabilities. \halt sweeps the verifier
match-logit margin. \full is the upper-right reference point.}
\label{fig:baseline-threshold-sweep-all3}
\end{figure*}

\paragraph{\halt threshold robustness.}
For \halt, Match is decided when
$\text{Match}_{\text{lp}}-\max(\text{Partial}_{\text{lp}},
\text{Null}_{\text{lp}})>\tau$. On 2Wiki, sweeping
$\tau \in \{-2,-1,0,1,2\}$ shifts the average stop loop from $1.85$
to $5.45$, while EM remains in $[0.182,0.196]$. The default
$\tau{=}0$ operating point is therefore inside a relatively flat region
of the EM curve. MuSiQue sweep statistics use the $885$ valid rows
(\S\ref{sec:appendix-stats}); the corresponding MuSiQue sweep is in
Table~\ref{tab:musique-threshold-sweep} and shows the same flat-EM
pattern between $\tau{=}0$ and $\tau{=}{+}2$.

\begin{table}[!ht]
\centering
\small
\setlength{\tabcolsep}{4pt}
\begin{tabular}{lrrrr}
\toprule
Method & $\tau$ & Fire & Avg.\ stop & EM \\
\midrule
\halt & $-2$   & 0.660 & 2.90 & 0.053 \\
\halt & $-1$   & 0.520 & 3.57 & 0.073 \\
\halt & $-0.5$ & 0.425 & 3.98 & 0.078 \\
\halt & $0$    & 0.298 & 4.44 & 0.086 \\
\halt & $+0.5$ & 0.162 & 4.96 & 0.087 \\
\halt & $+1$   & 0.075 & 5.29 & 0.088 \\
\halt & $+2$   & 0.016 & 5.50 & 0.087 \\
\bottomrule
\end{tabular}
\caption{\halt match-logit threshold sweep on MuSiQue, Self-Ask~3B
gold-claim condition, over the $n{=}885$ valid rows. Lowering $\tau$
makes the verifier fire earlier and saves loops at a steadily
increasing EM cost; the default $\tau{=}0$ sits at the knee.}
\label{tab:musique-threshold-sweep}
\end{table}

\paragraph{Sweep-vs-primary residual.} The $\tau$ sweep re-derives stop
decisions by re-thresholding cached per-claim match margins rather than
re-running the verifier end-to-end, so it does not reproduce the
primary pipeline's decisions exactly. At the default $\tau{=}0$ the
residual against Table~\ref{tab:primary} is at most $1.8$ points of
fire rate, $0.03$ loops, and $0.7\,$pp of EM across all three datasets:
$0.594$ vs $0.591$ fire and $0.195$ vs $0.199$ EM on 2Wiki, $0.658$ vs
$0.676$ and $0.290$ vs $0.297$ on HotpotQA (the largest gap), and
$0.298$ vs $0.294$ and $0.0859$ vs $0.0859$ on MuSiQue. We therefore
read the sweep as characterising the \emph{shape} of the loop--EM
frontier, with Table~\ref{tab:primary}'s $\tau{=}0$ values as the
reported operating point.

\subsection{Setting~B pilot details}
\label{sec:appendix-settingB-details}

The Setting~B pilot uses Self-Ask~3B over the FlashRAG Wikipedia 2018
corpus ($\sim 7.9$M passages) \citep{petroni2021kilt}, dense
E5-base-v2 retrieval \citep{wang2022e5}, and a FAISS index
(\S\ref{sec:appendix-settingB}). We focus on HotpotQA and Bamboogle
because both are sourced from English Wikipedia; using 2Wiki or
MuSiQue with this corpus would mix open-corpus retrieval failure with
snapshot mismatch.

\begin{table}[!ht]
\centering
\scriptsize
\setlength{\tabcolsep}{2pt}
\resizebox{\columnwidth}{!}{%
\begin{tabular}{lrrr}
\toprule
Metric & HotpotQA A ($n{=}1k$) & HotpotQA B ($n{=}1k$) & Bamboogle B ($n{=}125$) \\
\midrule
Native EM (agent)     & 0.292   & 0.115   & 0.112 \\
Full Std-Ext EM       & 0.306   & 0.109   & 0.072 \\
\halt \all EM         & 0.297   & 0.109   & 0.072 \\
$\Delta$EM (pp)       & $-0.9$  & $\phantom{+}0.0$ & $\phantom{+}0.0$ \\
Full loops            & 5.71    & 4.12    & 3.34 \\
\halt \all loops      & 3.15    & 3.96    & 3.31 \\
$\Delta$ loops        & $-45\%$ & $-4\%$  & $-0.7\%$ \\
Fire rate \all        & 0.676   & 0.064   & 0.024 \\
\bottomrule
\end{tabular}}
\caption{Setting~B pilot ($n{=}1000$ on HotpotQA, $n{=}125$ on
Bamboogle). HotpotQA B shows the effect of swapping the closed BM25
distractor pool for dense Wikipedia 2018 retrieval: underlying EM
drops sharply (0.292$\to$0.115), but \halt remains lossless
($\Delta$EM~$=0.0$~pp) while still trimming loops by $-4\%$.
The two Setting~B columns share the retrieval stack and chunking
configuration of \S\ref{sec:appendix-settingB}, but not their claim
source: HotpotQA~B is the gold-claim diagnostic condition, whereas
Bamboogle supplies no gold supporting-fact metadata and is necessarily
run with LLM-generated hop claims. The Bamboogle column is therefore
not a like-for-like comparison with the HotpotQA columns. Bamboogle
Setting~B passages are single sentences, so the per-chunk sentence cap
never binds (at most three qualifying sentences across all chunks).
On Bamboogle \halt fires on $2.4\%$ of questions, trimming loops by
$-0.7\%$ with $\Delta$EM~$=0.0$~pp.}
\label{tab:settingB}
\end{table}

\begin{figure*}[!t]
\centering
\includegraphics[width=\textwidth]{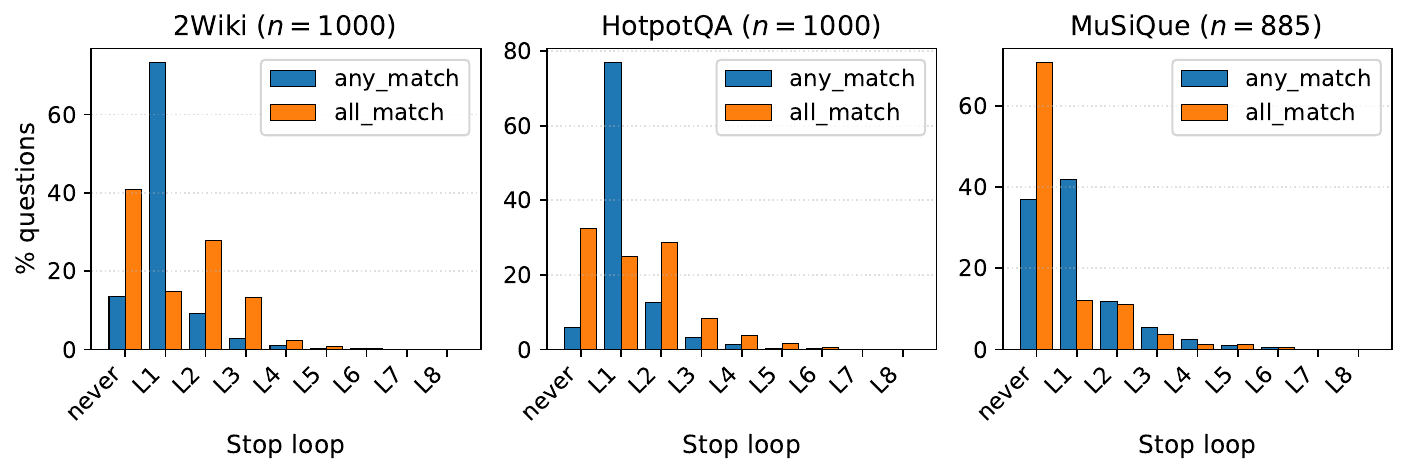}
\caption{Per-loop \halt fire rate on Self-Ask~3B. \textsc{all\_match}
spreads across early loops with a ``never'' tail, while
\textsc{any\_match} concentrates at loop~1. Each panel is over its
dataset's valid rows, so MuSiQue is over $885$
(\S\ref{sec:appendix-stats}); the ``never'' bin therefore counts only
questions the verifier declined to fire on, not questions with no
claims to verify.}
\label{fig:per-loop-fire}
\end{figure*}

The open-corpus pilot supports the conservative-firing interpretation.
On HotpotQA, replacing the closed distractor pool with dense Wikipedia
retrieval sharply lowers the underlying retrieval-and-answering
quality (Native EM~$0.292{\to}0.115$). The verifier responds by
firing on only $6.4\%$ of questions for \all (vs $67.6\%$ in
Setting~A), yet this conservative regime still trims average loops by
$-4\%$ with $\Delta$EM~$=0.0$~pp---i.e., \halt remains lossless under
open-corpus retrieval, simply because it abstains from stopping when
the verifier cannot confirm sufficiency.

\paragraph{Per-question case study (HotpotQA, Setting~B).}
We further audit the $64$~questions where \all fires. Among them, the
agent-extracted EM at the \halt-stop loop is $0.391$ versus $0.359$ at
the end of the full trajectory: stopping early is not merely safe but
$+3.1$~pp better on the fired subset. The selection mechanism is
sharply precision-oriented: fired-subset EM is $3.58\times$
$[2.55,4.78]$ the same-extractor population baseline of $0.109$
(paired-bootstrap 95\% CI, seed 13, 1000 iters; $\Delta\text{EM}{=}{+}28.2\,$pp
$[+16.3,+40.3]$), while the $936$ not-fired questions have a much lower
end-of-trajectory EM of $0.092$, showing \halt abstains exactly where
the agent is unlikely to be correct anyway. The more permissive \any
operating point ($18.4\%$ coverage) gives a smaller but still
significant lift, $+15.2\,$pp $[+9.4,+21.1]$ ($2.39\times$). The
per-question breakdown at the \all operating point gives $7$
wrong$\to$correct switches and $5$ correct$\to$wrong, yielding a net
$+0.20$~pp population EM shift that rounds to $0$ at three decimals. We
treat this as evidence that \halt's open-corpus behaviour is a
graceful-degradation feature rather than a limitation: under retrieval
distribution shift, the verifier defaults to abstention but retains a
usable, conservative firing surface that identifies the answerable
subset (verification-aware selective prediction) without a population
accuracy penalty. This Setting~B result uses \textbf{gold} hop claims;
we report it as a gold-claim diagnostic of open-corpus behavior, not as
a deployable open-corpus result.

Bamboogle ($n{=}125$, generated claims) reproduces the same selective
pattern on a smaller sample. \all fires on $3/125$ questions
($2.4\%$); on that fired subset EM is $0.333$ both at the \halt stop
loop and at the end of the full trajectory, and over the population EM
is unchanged ($0.072\to0.072$, F1 $0.1206\to0.1193$) while average
loops fall from $3.336$ to $3.312$. The more permissive \any operating
point fires on $13/125$ ($10.4\%$) and saves more search
($3.336\to3.168$ loops, $-5.0\%$), but costs EM on its fired subset
($0.308\to0.231$, one question lost) and over the population
($0.072\to0.064$) --- the same ordering between the two policies seen
in Setting~A. Loop savings are small in absolute terms because the
Bamboogle trajectories are short to begin with.

\subsection{Cost and latency profile}
\label{sec:appendix-cost}

\halt adds sentence-level verifier calls and avoids generator search
loops. Table~\ref{tab:cost} reports the verifier-side wall-clock in
our implementation; Table~\ref{tab:wallclock} below reports measured
end-to-end wall-clock, and Table~\ref{tab:token-accounting} reports the
corrected raw-token accounting.

\begin{table}[!ht]
\centering
\footnotesize
\setlength{\tabcolsep}{3pt}
\resizebox{\columnwidth}{!}{%
\begin{tabular}{lrrr}
\toprule
Setting & $\Delta$loops & Ver.\ calls/q & Ver.\ time$^\ast$ \\
\midrule
Self-Ask 3B 2Wiki    & $-2.05$ &  9.7  & $4.16\,$s \\
Self-Ask 3B HotpotQA & $-2.56$ &  8.6  & $3.81\,$s \\
Self-Ask 3B MuSiQue  & $-1.09$ & 12.03 & $5.92\,$s \\
\bottomrule
\end{tabular}}
\caption{Verifier-side cost profile for \all at 3B scale, gold-claim
condition. $^\ast$\,Qwen2.5-3B + LoRA, unbatched wall-clock.}
\label{tab:cost}
\end{table}

\paragraph{Raw token accounting is net positive for \halt.}
A na\"ive token count is the wrong cost metric, but we report it for
completeness. Using the exact \texttt{Qwen2.5-3B-Instruct} tokenizer
against the cached agent transcripts, Table~\ref{tab:token-accounting}
gives the corrected six-arm accounting (3 datasets $\times$
gold/generated claims). \halt's verifier calls process more raw tokens
than the generator decoding they avoid, on every arm: net raw
tokens/query range from $+4{,}993$ (HotpotQA gold) to $+8{,}503$
(MuSiQue generated). We state this plainly: \emph{\halt's efficiency
claim is about search iterations and measured wall-clock, not raw
token count.} The verifier's tokens are cheap parallel-prefill
scoring calls, while the generator tokens it avoids are expensive
sequential decode steps under repeated full re-prefill; the batched
per-query latency (rightmost column, at a measured $116\,$ms per
batch-8 claim--evidence pair) remains favorable throughout.

\begin{table*}[!ht]
\centering
\scriptsize
\setlength{\tabcolsep}{3pt}
\resizebox{\textwidth}{!}{%
\begin{tabular}{llrrrrrrr}
\toprule
Dataset & Claims & $n$ & $\Delta$loops & Ver.\ calls/q & Gen.\ out.\ tok saved/q & Ver.\ tok processed/q & Net raw tok/q (ver $-$ saved) & Batched latency/q \\
\midrule
HotpotQA  & gold & 1000 & $-2.56$ &  8.58 & 98 & 5{,}092 & $+4{,}993$ & $1.00\,$s \\
HotpotQA  & gen  & 1000 & $-1.01$ & 10.98 & 39 & 6{,}518 & $+6{,}479$ & $1.27\,$s \\
2WikiMQA  & gold & 1000 & $-2.05$ &  9.66 & 66 & 5{,}735 & $+5{,}669$ & $1.12\,$s \\
2WikiMQA  & gen  & 1000 & $-0.95$ & 10.31 & 31 & 6{,}122 & $+6{,}092$ & $1.20\,$s \\
MuSiQue   & gold &  885 & $-1.09$ & 12.03 & 41 & 7{,}142 & $+7{,}101$ & $1.40\,$s \\
MuSiQue   & gen  & 1000 & $-0.11$ & 14.33 &  4 & 8{,}507 & $+8{,}503$ & $1.66\,$s \\
\bottomrule
\end{tabular}}
\caption{\textbf{Corrected six-arm token accounting.} Generator
output tokens/loop extracted verbatim from persisted transcripts
(Qwen2.5-3B tokenizer, exact match against \texttt{tool\_trace} length
on 5{,}705/5{,}517/5{,}516 loops for hotpot/2wiki/musique). Verifier
prompt length uses the measured mean $197.86$ tokens/call ($n{=}128$
distinct prompts, min $171$/max $246$), $\times 3$ for the 3-way label
forward pass. MuSiQue gold is reported over its $885$ valid rows
(115/1000 rows have no atomic claims to verify and are excluded from
both numerator and denominator; see \S\ref{sec:appendix-cost} note
below). Net raw tok/q is positive on every arm: \halt costs more raw
tokens than it saves. Batched latency/q is calls/q $\times\,116\,$ms
(measured batch-8 ms/pair).}
\label{tab:token-accounting}
\end{table*}

\paragraph{MuSiQue skip-row correction.} MuSiQue gold has $115/1000$
rows with no atomic claims to verify (\texttt{skip\_reason:
no\_claims}), which carry no loop or verifier-call fields. Averaging
by dividing the sum over the $885$ valid rows by the full $n{=}1000$
silently treats the $115$ skip rows as zero-loop, zero-verifier-call
rows in the denominator, biasing every MuSiQue gold average low; we
use the valid-row denominator ($n{=}885$) throughout this appendix and
in Table~\ref{tab:primary}.

\paragraph{Measured end-to-end wall-clock.} Table~\ref{tab:cost}'s
verifier-time column and Table~\ref{tab:token-accounting}'s
batched-latency column isolate verifier-side cost, but the actual
question is whether \halt saves wall-clock once the avoided generator
loops are accounted for. We measure this directly rather than
estimating it. Using timed replay decomposition on HotpotQA
($n{=}100$, Setting~A, Self-Ask 3B, greedy decoding so loop counts are
deterministic and match the cached trajectories $100/100$), we run the
agent live with per-loop timestamps, apply the cached \halt stop
loops, reconstruct claim--evidence pairs offline (validated $100/100$
against cached verifier calls), and score them live with the batched
($B{\leq}8$) verifier.

\begin{table}[!ht]
\centering
\footnotesize
\setlength{\tabcolsep}{3pt}
\resizebox{\columnwidth}{!}{%
\begin{tabular}{lrrrr}
\toprule
Arm & Mean s/q & Mean loops & Ver.\ share & $\Delta$ vs \full [95\% CI] \\
\midrule
\full & 47.40 & 5.80 & --- & --- \\
\halt-gold & 30.94 & 3.49 & 6.5\% & $-16.45\,$s ($-35\%$) $[-19.9,-12.9]$ \\
\halt-gen (deployable) & 40.62 & 4.75 & 5.4\% & $-6.77\,$s ($-14\%$) $[-10.2,-3.3]$ \\
\bottomrule
\end{tabular}}
\caption{\textbf{Measured end-to-end wall-clock}, HotpotQA $n{=}100$,
paired bootstrap (seed $13$, $1000$ iters). Both \halt arms save
wall-clock with 95\% CIs strictly below zero. The verifier is only
$5$--$7\%$ of total time; each avoided loop costs
${\approx}16.45/2.31{\approx}7.1\,$s of wall-clock, because every loop
issues three \texttt{.generate()} calls with full re-prefill on the
4-bit model (retrieval is ${<}0.3\%$ of loop time). This measured rate
supersedes an earlier unmeasured ``$1.0$--$1.5\,$s per loop'' estimate
that had propagated into a mislabeled ``per avoided loop'' figure; we
retire that estimate in favor of the measured numbers above.}
\label{tab:wallclock}
\end{table}

\subsection{Per-loop fire-rate distribution}
\label{sec:appendix-extended-fire}

On 2Wiki, \all firings spread over loops 1--3 with a long ``never''
tail (41\%). HotpotQA is similarly multi-modal (L1 25\%, L2 29\%,
never 32\%). MuSiQue has a lower overall fire rate ($0.294$ over its
$885$ valid rows, versus $0.591$ and $0.676$), but its fired cases
still span loops 1--3. By contrast, \any fires at loop~1 on $73\%$ of
2Wiki, $77\%$ of HotpotQA and $42\%$ of MuSiQue questions.

\subsection{When does \halt help?}
\label{sec:appendix-when-help}

\paragraph{Controlled distractor dose.}
As a diagnostic for distractor contamination, we vary the number of
distractor passages per question on HotpotQA at $n{=}200$. At 1, 5,
and 7 distractors, $\Delta$EM for \halt \all minus \full is
$+0.5$, $+0.5$, and $+2.0\,$pp. The direction is consistent with the
hypothesis that avoiding extra post-coverage context can help, but the
absolute effect is small. We therefore treat this as secondary
diagnostic evidence rather than as the main mechanism argument.

\paragraph{MuSiQue floor subset.}
On MuSiQue, the low absolute EM baseline makes aggregate changes small.
We split questions by whether \full is correct under the standardized
extractor. In the \full-correct subset ($n{=}73$), \all fires on 51\%
of questions with safe-stop rate 84\%. In the \full-incorrect subset
($n{=}927$), \all fires on 27\% of questions with safe-stop rate 96\%.
The overall $\Delta\text{EM}{=}{+}0.3\,$pp is therefore the net effect
of a small number of correct$\to$incorrect and incorrect$\to$correct
changes, rather than a systematic floor-effect artifact.

\subsection{$K$ sensitivity and lexical-overlap checks}
\label{sec:appendix-K-sens}

Truncating to a single hop claim ($K{=}1$) drops the \all fire rate
from $0.591$ to $0.493$ on 2Wiki, from $0.676$ to $0.395$ on HotpotQA,
and from $0.260$ to $0.133$ on MuSiQue. The drop is larger on datasets
with longer or more decisive later-hop structure. This supports the
main analysis: second-and-later hop claims affect whether \halt fires,
not merely when it fires.

\subsection{Stronger-extractor robustness: supplementary detail}
\label{sec:appendix-stronger-extractor}

Table~\ref{tab:stronger-extractor} reports the main stronger-extractor
result. Two additional observations are useful. First, the standardized
extractor is important for comparing adaptive controls: on a 2Wiki
$n{=}200$ diagnostic, FLARE has native EM $0.000$ but Std-Ext EM
$0.135$, while DRAGIN-approx has native EM $0.000$ and Std-Ext EM
$0.145$. Second, the HotpotQA $\Delta$EM under the 7B extractor narrows
from $-2.2\,$pp at $n{=}500$ to $-0.7\,$pp at $n{=}1000$, indicating
that the earlier larger gap was within sampling variability.

\subsection{Verifier ablation details}
\label{sec:appendix-ablations}

Table~\ref{tab:verifier-ablations} reports the three controlled
ablations summarized in \S\ref{sec:verifier-ablations}, run on the same
Self-Ask~3B trajectories.

\paragraph{(a) Random hop claim.} We replace each question's expected
hop claims with claims from a different question. The \all fire rate
collapses ($0.591 \to 0.034$ on 2Wiki, $0.676 \to 0.196$ on HotpotQA,
$0.260 \to 0.018$ on MuSiQue): the verifier rarely fires when the
expected claim is unrelated to the retrieved evidence.

\paragraph{(b) Page-fixed claim-random.} The random-claim ablation
changes both the claim text and the \textsc{Expected page} field, and
dropping \textsc{Expected page} alone reduces verifier macro-F1 by
$0.176$ (\S\ref{sec:appendix-verifier}), so the collapse could partly
come from page-title mismatch. Keeping the \textsc{Expected page} field
fixed to the question's own gold supporting-fact title while replacing
the other claim fields recovers only a small part of the firing
($0.034 \to 0.105$ on 2Wiki, $0.196 \to 0.216$ on HotpotQA,
$0.018 \to 0.017$ on MuSiQue), far below the gold baselines: page
information contributes, but does not explain the stopping signal.

\paragraph{(c) Evidence word-shuffle.} Shuffling word order within each
retrieved chunk keeps the same tokens but disrupts sentence-level
meaning; firing collapses ($0.591 \to 0.049$ / $0.676 \to 0.077$ /
$0.260 \to 0.008$), so the verifier is not detecting bag-of-words
overlap.

\begin{table}[t]
\centering
\footnotesize
\setlength{\tabcolsep}{3pt}
\begin{tabular}{lrrr}
\toprule
Condition & 2Wiki & HotpotQA & MuSiQue \\
\midrule
Baseline (gold)                    & 0.591          & 0.676          & 0.260          \\
(a) Random hop claim               & \textbf{0.034} & 0.196          & \textbf{0.018} \\
(b) Page-fixed claim-rand          & 0.105          & 0.216          & 0.017          \\
(c) Evidence word-shuffle          & 0.049          & 0.077          & 0.008          \\
\bottomrule
\end{tabular}
\caption{\textbf{Verifier ablations on Self-Ask~3B trajectories.}
Each cell shows the \all fire rate, i.e., the fraction of questions on
which \halt halts. Rows~(a)--(c) test claim mismatch, page-title
control, and word-order disruption.}
\label{tab:verifier-ablations}
\end{table}

\subsection{Bidirectional-control protocol details}
\label{sec:appendix-continue}

\paragraph{Protocol.} The forced-continuation runs in
\S\ref{sec:results-continue} re-run the 7B agent on the flagged
under-covered subset with the released \texttt{--force-continue} flag:
the agent's native ``no follow-up'' decision is overridden and search
continues loop-by-loop until \all fires or the budget (8 loops) is
exhausted; \halt then stops each forced trajectory at its own \all
loop. This is the same ``agent unchanged, verifier controls the loop''
philosophy applied to extend rather than cut: \halt supplies only the
continuation signal, never search queries or answer content.

\paragraph{Mechanism-control subset sizes.} Splits below are recovered\,/\,non-recovered and
sum to the forced subset size ($511$ HotpotQA, $635$ 2Wiki, $723$
MuSiQue). Under gold claims, coverage recovers under forcing for
$290/221$ of the HotpotQA under-covered questions ($+26.2$\,pp EM for
the recovered group versus $+2.7$\,pp for the rest), $311/324$ on
2Wiki, and $135/588$ on MuSiQue (i.e.\ only $19\%$ of forced MuSiQue
trajectories reach coverage). Recovery is markedly lower when hop
claims are generated from the question instead of taken from gold
metadata: $114/397$ on HotpotQA, $182/453$ on 2Wiki and $14/709$ on
MuSiQue, consistent with noisier generated claims being harder to
satisfy even under forced continuation. In every condition the
non-recovered majority stays near its native-stop EM --- forcing
continuation on trajectories that never reach coverage barely helps.

\paragraph{MuSiQue de-duplication.} The MuSiQue $n{=}1000$ slice holds
only $911$ unique question ids: $89$ questions were scored twice.
Flagging under-covered trajectories by raw row therefore gives $793$
rows, which collapse to $723$ unique questions ($60$ duplicated
under-covered and $10$ duplicated no-claims rows; the other $19$
duplicates fall on already-covered questions). The forced-continuation
artifact is exactly that $723$-question set (both symmetric differences
against the flagged set are empty), so we report $n{=}723$. This is a
duplication in the sampled dev slice, not a modelling artifact.

\paragraph{Extractor consistency.} Native-stop and forced-continuation
EM, and the selective-continuation metric that combines them, are all
read from the same standardized 3B-extractor field
(\texttt{policies.\{full,all\_match\}.em}) rather than from either
agent's own-answer proxy.

\end{document}